%% file: main.tex
\documentclass{article}

\usepackage[preprint]{neurips_2026}

\usepackage[utf8]{inputenc}
\usepackage[T1]{fontenc}
\usepackage{microtype}
\usepackage{url}
\usepackage[blocks]{authblk}
\usepackage{amsmath,amssymb,amsfonts,amsthm,mathtools,bm}

\usepackage[pdftex]{graphicx}
\usepackage{xcolor}
\usepackage{booktabs}
\usepackage{multirow}
\usepackage{makecell}
\usepackage{tabularx}
\usepackage{array}
\usepackage{caption}
\usepackage{subcaption}
\usepackage{wrapfig}
\usepackage{paracol}
\usepackage{placeins}
\usepackage{nicefrac}

\usepackage{soul}

\usepackage[linesnumbered,ruled,lined,noend,vlined]{algorithm2e}
\usepackage{algpseudocode}
\usepackage{titlesec}

\usepackage{hyperref}

\input{math_commands.tex}

\title{LIGO-PINN: Learned Initialization via Gated Optimization to Alleviate Convergence Failures in Physics Informed Neural Networks}

\author[1]{Nilay Anurag}
\author[1,*]{Shital Adhikari}
\author[2]{Taniya Kapoor}
\author[1,*]{Nikhil Muralidhar\vspace{1.2ex}}

\affil[1]{\normalfont Department of Computer Science, Stevens Institute of Technology}
\affil[2]{\normalfont Artificial Intelligence Group, Wageningen University \& Research\\\vspace{1.2ex}}

\affil[ ]{\normalfont\texttt{\{nanurag, sadhikar1, nmurali1\}@stevens.edu, taniya.kapoor@wur.nl}}

\date{} 

\newcommand{\red}[1]{\textcolor{red}{#1}}
\newcommand{\hide}[1]{}

\newcommand{\methodcurriculumn}{Curr-Reg.}
\newcommand{\methodevo}{R3}
\newcommand{\methodpinnd}{PINN (dynamic)}
\newcommand{\methodpinnf}{PINN (fixed)}
\newcommand{\methodmetapde}{Meta-PDE}
\newcommand{\methodmaml}{LIGO-PINN w/o (GLO)}
\newcommand{\methodhyperlr}{HyperLR}
\newcommand{\mymethod}{LIGO-PINN}

\newcommand{\mymethodfull}{Physics Informed Neural Network with Learned Initialization via Gated Optimization}

\begin{document}

\begingroup
\let\thefootnote\relax\footnotetext{*Corresponding Authors: \href{sadhikar1@stevens.edu}{sadhikar1@stevens.edu}, \href{nmurali1@stevens.edu}{nmurali1@stevens.edu}}
\endgroup

\maketitle
\vspace{-3ex}
\begin{abstract}
\vspace{-2ex}
Physics-informed neural networks (PINNs) have had a broad research impact in modeling domains governed by partial differential equations (PDE). 
\hide{However, PINNs perform poorly, when generalizing to unseen but related PDE domains. Further, PINNs converge to trivial solutions, when modeling challenging PDE domains.} However, PINNs have been shown to perform poorly, sometimes even converging to trivial solutions, in challenging PDE domains, or when generalizing to unseen but related PDE domains. Previously proposed solutions to alleviate this well known PINN failure mode, detail hyperparameter tuning to reduce loss imbalance between data-driven and physics guided losses, curriculum learning based training strategies or ones involving dynamic re-sampling of \emph{hard} collocation points. We observe that these methods face certain pitfalls e.g., hyperparameter tuning is expensive, designing a training curriculum is ambiguous in multi-parameter PDE settings and dynamic resampling of collocation points still fails in complex PDE settings.\hide{Yet other work has highlighted that conflicting gradients during PINN training, may be one of the major influences of such catastrophic PINN failures.}
Complementary to this line of thinking, we believe the initial PINN network weights also play a crucial role in the emergence of catastrophic failures during training. However, the effect of PINN weight initialization has been surprisingly under-investigated. To this end, we propose a novel PINN training methodology based on \emph{Learned Initialization}, to address catastrophic PINN failures. Specifically, we propose a framework for \emph{Learned Initialization via Gated Layerwise Optimization} (\mymethod{}) to overcome PINN convergence failures.\hide{when training on challenging PDE domains.} Through rigorous evaluation on 1D and 2D PDE domains, including a challenging 2D fluid dynamics setting, we demonstrate that our proposed methodology outperforms state-of-the-art methods, including those specifically designed to alleviate PINN failures. We show that \mymethod{} achieves a \textbf{91.5\%} average performance improvement across six state-of-the-art PINN baselines and \textbf{81\%} performance improvement over the strongest baseline, when evaluated across three diverse PDE domains. We also verify that \mymethod{} generalizes to 3D unstructured domains.
Finally, we conduct a rigorous analysis to identify and explain the performance improvement of \mymethod{} and the convergence failure of traditional PINNs, by analyzing training dynamics across all three PDE domains. Our code is publicly available at \href{https://github.com/scailab/ligo-pinn}{https://github.com/scailab/ligo-pinn}.
\hide{We complement the empirical analysis by also undertaking a rigorous training dynamics characterization to analyze the reasons for catastrophic PINN failures by investigating the PINN loss dynamics, and spectral bias.} 
\hide{Finally, we also demonstrate that our proposed LeIn-PINN method significantly reduces spectral bias compared to traditional PINNs even in challenging PDE domains.}
\end{abstract}

\input{sections/c1.Introduction/intro}
\input{sections/c2.RelatedWork/related}

\input{sections/c3.theory/theory}
\input{sections/c4.ProblemFormulation/problem}

\input{sections/c4.ProblemFormulation/dataset}
\input{sections/c6.ResultsAnalysis/results_and_discussion}

\input{sections/c7.Conclusion/summary}

\clearpage

{\small
\bibliographystyle{plainnat}
\bibliography{sections/ref}
}
\clearpage
\input{sections/C9.Appendix/appendix}
\end{document}

%% file: math_commands.tex
\def\eqref#1{equation~\ref{#1}}

\def\1{\bm{1}}

\DeclareMathAlphabet{\mathsfit}{\encodingdefault}{\sfdefault}{m}{sl}
\SetMathAlphabet{\mathsfit}{bold}{\encodingdefault}{\sfdefault}{bx}{n}



%% file: sections/c1.Introduction/intro.tex
\section{Introduction}
\vspace{-1.5ex}
Partial differential equations (PDEs) govern dynamics in heat transfer, fluid flow, electromagnetics, and other physical systems. Efficient PDE solvers are central to scientific progress. Physics-Informed Neural Networks (PINNs)~\citep{raissi2017physicspart1,raissi2018numerical} embed PDE constraints into neural network loss functions, enabling both forward and inverse problem solving. PINNs have been applied widely~\citep{cuomo_scientific_2022}, but often fail in \emph{hard} PDE settings, converging to trivial or inaccurate solutions. A key challenge is the imbalance between data-driven and physics-informed loss terms~\citep{wang2021understanding}, which causes optimization pathologies. Remedies include curriculum learning~\citep{krishnapriyan_characterizing_2021}, adaptive sampling~\citep{daw_mitigating_2023}, and spectral/Neural Tangent Kernel (NTK) analyses~\citep{wang2021eigenvector,wang2022and,lau2024pinnacle}. Yet these approaches typically succeed only in narrow PDE regimes and lack generality. Meanwhile, the machine learning community has shown that initial weights strongly influence trainability in deep networks~\citep{arpit2019initialize,schoenholz_deep_2017,taki2017deep,yang2017mean}. Standard weight initialization schemes such as Xavier or Kaiming can place networks in chaotic or overly ordered regimes, leading to poor convergence. Despite this, nearly all PINN studies rely on off-the-shelf initializers without task-specific adaptation.

To address this gap, we introduce \emph{\textbf{L}earned-\textbf{I}nitialization via \textbf{G}ated Layerwise \textbf{O}ptimization} for PINNs (\textbf{\mymethod{}}), a framework to systematically initialize PINN network weights that alleviates catastrophic training dynamics. Our contributions are as follows:
\begin{itemize}
    \item We propose a method to systematically \emph{learn} the initial PINN weights, to mitigate catastrophic training failures, commonly observed in PINNs, particularly in challenging PDE regimes.
    \item As part of the \emph{learned initialization} method, we introduce a novel gated layer-wise optimization (GLO) procedure and show that it produces initial PINN weights that mitigate spectral bias during PINN training.
    \item Through extensive experiments on three diverse PDE domains\hide{1D Convection, 2D Helmholtz, and 2D Navier--Stokes,}, we demonstrate consistent improvements over state-of-the-art baselines, especially in extrapolation settings.
\end{itemize}

%% file: sections/c2.RelatedWork/related.tex
\section{Related Work}
We now review major research directions investigating PINN training failures~\citep{cuomo2022scientific}.\hide{Here we review the most recent, representative work.}
\vspace{-1ex}
\par\noindent
\textbf{Temporal Dynamics, Sampling, and Curriculum-Based Methods.}
A seminal work  ~\citep{wang2021understanding} uncovered the imbalance in gradients between data-driven and physics-based loss terms as one of the reasons for catastrophic PINN failures, causing converge to trivial solutions. Early work also highlighted the importance of respecting causal order of PDE dynamics during training, in time-dependent problems~\citep{wang2022respecting}.  Building on this, residual-based sampling techniques dynamically reweigh collocation points in regions with high PDE residuals~\citep{daw_mitigating_2023,toloubidokhti2023dats}.  In parallel, curriculum learning approaches schedule training tasks to be ordered by increased difficulty, guiding PINNs from simpler to complex PDE regimes~\citep{krishnapriyan_characterizing_2021}.  While these strategies reduce domain errors, they require careful tuning when PDE parameters vary widely and do not generalize well across different PDE systems.
\vspace{-1ex}
\par\noindent
\textbf{Gradient-Level Interventions}
Efforts to address the issue of loss imbalance in PINNs have also explored gradient-level solutions. \cite{kim2021dpm} introduced a Dynamic Pulling Method (DPM) which employs a pseudo-inverse operation to align losses in the same direction, to improve stability, effectiveness of training. \cite{wang2025gradientalignmentphysicsinformedneural} investigated effect of higher-order optimizers.
\vspace{-1ex}
\par\noindent
\textbf{Meta-Learning for PINNs.}
Several recent works have explored meta learning concepts in PDE systems.
Psaros et al.~\citep{psaros_meta-learning_2022} uses meta learning to discover optimal loss function for different PINN system. In contrast  \mymethod{} uses meta learning to inform PINN weight initialization.
Hyper-LR-PINNs~\citep{cho2023hypernetwork} employ a hyper-network architecture to generate low-rank, task-specific weights for each PDE domain. Although they have trained on similar PDE domains to ours, their experiments have not explored extrapolation regimes as extreme as ours in this paper. 
\cite{qin2022metapdelearningsolvepdes} is the closest paper to ours and also employs a model-agnostic meta-learning (MAML) approach for PINN training. However, their evaluation is restricted to simple PDE configurations and unlike our setup, they do not investigate gated optimization strategies or generalization to unseen PDE configurations.
\vspace{-1ex}
\par\noindent
In contrast to other approaches to alleviate catastrophic training failures in PINN,~\mymethod{} is the only approach to systematically explore the effect of learned weight initialization on PINN training dynamics. Further, our gated layer-wise optimization is the first approach of its kind applied in the context of addressing convergence failures in PINNs across diverse PDE settings.

%% file: sections/c4.ProblemFormulation/problem.tex
\section{Problem Formulation}\label{sec:lein_pinn_formulation}
\vspace{-1.5ex}
We follow the standard PINN setting as in~\citep{raissi2017physicsinformeddeeplearning}, also described briefly below; See Appendix~\ref{appendix:pinn_formulation} for detailed formulation.
Let $\mathbf{z} \in \Omega \subset \mathbb{R}^{d}$ denote the spatial coordinate within domain $\Omega$ (with boundary $\partial\Omega$), $t \in [0,T]$ time, and $u(\mathbf{z},t)$ the physical state solving the PDE in Eq.~\ref{eq:pde_formulation}.
\begin{equation}
\mathcal{F}\big(u(\mathbf{z},t);\gamma\big) = f(\mathbf{z},t),
\quad \mathbf{z} \in \Omega,\ t \in [0,T],
\qquad
\mathcal{B}\big(u(\mathbf{z},t)\big)=u_b(\mathbf{z},t),\ \mathbf{z} \in \partial\Omega,
\label{eq:pde_formulation}
\end{equation}
Here, $\mathcal{F}$ is the (possibly nonlinear) differential operator representing the underlying physical law, $\gamma$ are the PDE parameters (e.g., viscosity, diffusivity), and $f(\mathbf{z},t)$ is a known forcing or source term.
The boundary operator $\mathcal{B}$ enforces boundary or initial conditions, and $u_b(\mathbf{z},t)$ specifies the prescribed values along $\partial\Omega$ and the subscript `$b$' is added to indicate solution along the boundary $\partial\Omega$.

A neural network $\hat u_\Theta(\mathbf{z},t)$ parameterized by weights $\Theta$ is trained to satisfy the PDE at interior (collocation) points and to match boundary/initial data on $\partial\Omega$.
With $N_{\Omega}$ interior samples ${(\mathbf{z}_i,t_i)}$ and $N_{\partial\Omega}$ boundary/initial samples ${(\mathbf{z}_j,t_j)}$, the objective is listed in Eq.~\ref{eq:pinn-total-loss}.
\vspace{-1ex}
\begin{equation}
  \label{eq:pinn-total-loss}
  \mathcal{L}(\Theta)
  = \lambda_r \frac{1}{N_{\Omega}} \sum_{i=1}^{N_{\Omega}}
     \Big|\mathcal{F}(\hat u_\Theta(\mathbf{z}_i,t_i);\gamma) - f(\mathbf{z}_i,t_i)\Big|^2
   + \frac{1}{N_{\partial\Omega}} \sum_{j=1}^{N_{\partial\Omega}}
     \Big|\hat u_\Theta(\mathbf{z}_j,t_j) - u_b(\mathbf{z}_j,t_j)\Big|^2.
\end{equation}
Here, $\lambda_r$ balances the contributions of the physics residual and boundary losses.

\hide{We study three canonical PDE systems commonly used to benchmark scientific machine learning: 1D Convection, the 2D Helmholtz equation, and the 2D incompressible Navier--Stokes equations. Each captures a distinct physical regime—transport, wave propagation, and nonlinear fluid dynamics—that poses different challenges for PINN optimization.}




\hide{Although the PINN formulation has been highly impactful in solving \emph{forward} problems across variegated PDE  domains, it is well known that PINNs fail to learn faithful solutions in challenging (e.g., stiff) PDE contexts.} Previous work has tried to alleviate PINN training failures by hyper-parameter tuning of $\lambda_r$~\cite{wang2021understanding}, learning improved training curricula~\cite{krishnapriyan_characterizing_2021} or by intelligently re-sampling collocation points to focus on \emph{hard} regions of the domain~\cite{daw_mitigating_2023,wu2023comprehensive} but has all seen limited success as our investigations of these methods in Sec.~\ref{sec:results} reveal. 
\hide{Previous work~\citep{wang2021understanding} has characterized the conflicting and imbalanced gradients between the two loss terms in $\mathcal{L}(\Theta)$ as one of the reasons for such failures. Recent approaches  have proposed solutions to address PINN convergence failures by R3 re-sampling ~\citep{daw_mitigating_2023,wu2023comprehensive}, curriculum based PINN training~\citep{krishnapriyan_characterizing_2021}.}

Complementary to these works, we hypothesize that PINN convergence failures in \emph{challenging} (a.k.a. hard) PDE domains, can be alleviated, if PINN weights $\Theta$, instead of being randomly initialized, are systematically initialized prior to training on the \emph{hard} domain. To this end, we propose a \emph{Learned-Initialization} mechanism to systematically learn better initial weights, such that PINNs trained with these learned initial weights overcome catastrophic training failures. Our proposed \mymethod{} method, has two facets (i) Invariance Encoding (ii) Gated Layer-wise Optimization. 

\begin{wrapfigure}{r}{0.58\textwidth} 
\small
\vspace{-4ex}
\begin{algorithm}[H]
  \small
  \caption{\mymethodfull{}}
  \label{alg:mlayer_iif}
  \LinesNumberedHidden
  \DontPrintSemicolon
  \KwIn{Tasks \(p(\Gamma_{\mathrm{easy}})\), initial weights \(\Theta_0\), inner LR \(\eta_{\mathrm{inner}}\), outer LR \(\eta_{\mathrm{outer}}\),\\
         total meta-iterations \(\mathcal{R}\), Num. Sampled Tasks \(k\)}
  \KwOut{Learned Initial Weights \(\Theta_\mathcal{R}\)}
  \For(\tcp*[h]{meta-iterations \(r=1,\dots,\mathcal{R}\)}){\(r\leftarrow0\)}{
     Sample batch \(\{\mathcal{T}_i\}_{i=1}^k\sim p(\Gamma_{\mathrm{easy}})\)\;
     Copy Global Weights \(\{\Theta'_{r,i}\}_{i=1}^k:= \Theta_r\)\;
    \For{\(i=1\) \KwTo \(k\)}{
       \(\Theta''_{r,i}\gets\Theta'_{r,i} - \eta_{\mathrm{inner}}\nabla_{\Theta'_{r,i}}\mathcal{L}_{T_i}(\Theta'_{r,i})\)\;
    }
     Compute Meta Gradients \([\nabla_{\theta^0_r}\mathcal{L}_{\mathrm{meta}},\dots,\nabla_{\theta^{L-1}_r}\mathcal{L}_{\mathrm{meta}}]\)\;
    \For{\(l=0\) \KwTo \(L-1\)}{
       \(g^l(r)=\mathbf{1}\{l\le\lfloor \frac{rL}{\mathcal{R}}\rfloor\}\)\;
       \(\theta^{l}_{r+1}\gets\theta^{l}_r - \eta_{\mathrm{outer}}g^l(r)\nabla_{\theta^{l}_r}\mathcal{L}_{\mathrm{meta}}\)\;
    }
  }
\end{algorithm}
\end{wrapfigure}
\vspace{-0.5ex}
\textbf{Invariance Encoding (IE).} If the goal of the PINN is to learn a challenging PDE domain governed by PDE parameter $\gamma_{\mathrm{hard}}$, IE seeks to first train a randomly initialized PINN model on a set of \emph{relatively easier} configurations of the PDE with parameters $\Gamma_{\mathrm{easy}} = \{\gamma_{1},\dots,\gamma_{K}\}$. This enables a distillation of the \emph{invariant} physics across $\Gamma_{\mathrm{easy}}$ into the PINN weights $\Theta$. Our approach for IE is grounded in  foundational works of meta‐learning~\citep{finn_model-agnostic_2017}.  IE begins with randomly initialized \emph{global} PINN weights \(\Theta_0\) and a pool of $K$ tasks in $\Gamma_{\mathrm{easy}}$.
The IE learning process is carried out for $\mathcal{R}$ iterations $\{1,2,\dots,\mathcal{R}\}$ and at each training iteration $r$, tasks $\{\mathcal{T}_1,\dots,\mathcal{T}_k\}$ are randomly sampled from the task distribution $p(\Gamma_{\mathrm{easy}})$ of easy tasks and $k$ identical copies $\{\Theta'_{r,1},\dots,\Theta'_{r,k}\}$ of the current global PINN weights $\Theta_r$ are created. Each $\Theta'_{r,i}$ is then optimized via. gradient descent, employing data from task $\mathcal{T}_i$ at iteration $r$ for `J' \emph{inner} iterations (for simplicity we assume `J' = 1). Let $\Theta''_{r,i}$ represent the updated version of $\Theta'_{r,i}$ after `J' inner gradient-descent update iterations.  
The \emph{common} representations across all $k$  task models $\{\Theta''_{r,1},\dots,\Theta''_{r,k}\}$ is distilled into $\Theta_r$ via. a `meta-update' as indicated in Eq.~\ref{eq:meta_update} where $\mathcal{L}_{\mathcal{T}_i}(\Theta''_{r,i})$ represents the loss in Eq.~\ref{eq:pinn-total-loss} calculated w.r.t parameters $\Theta''_{r,i}$ with data from task $\mathcal{T}_i$ and $\eta_{\mathrm{outer}}$ represents the learning rate.
\vspace{-2ex}
\begin{align}
\Theta_{r+1}  = \Theta_{r} - \eta_{\mathrm{outer}}\nabla_{\Theta_r}\mathcal{L}_{\mathrm{meta}};\qquad\qquad
\nabla_{\Theta_r}\mathcal{L}_{\mathrm{meta}}  =\nabla_{\Theta_r}\sum_{i=1}^k\mathcal{L}_{\mathcal{T}_i}(\Theta''_{r,i})
\vspace{-2ex}
\label{eq:meta_update}
\end{align}
\noindent\textbf{Gated Layer-wise Optimization (GLO).} Traditional meta-learning approaches~\citep{finn_model-agnostic_2017,rajeswaran2019meta} are not physics-informed and treat all tasks in $\Gamma_{\mathrm{easy}}$ as resulting in similar training dynamics across all layers. Thus, the meta-update in Eq.~\ref{eq:meta_update} is a simple summation of all task gradients. However, based on previous work~\citep{wang2021understanding}, we know that non-trivial dynamics between the physics-based  loss and data-driven loss affects different layers in the neural network differently. To address these gaps that manifest uniquely in the physics-informed PDE modeling context, we augment the IE procedure with the novel GLO mechanism by introducing a \emph{gated} update of layers during each meta-update, such that a gating mechanism $g(r)$, \emph{unlocks} deeper layers, exposing their parameters to gradient-based optimization, gradually, over the training iterations from 1 to $\mathcal{R}$.

Let us consider that the global PINN is an `L' layer neural network whose parameters at iteration $r$ are represented as $\Theta_r = \{\theta_r^0,\dots,\theta_r^{L-1}\}$ with $\theta^i_r$ representing the weights corresponding to layer $i$ at iteration $r$. \hide{Standard IE via. meta-learning~\citep{finn_model-agnostic_2017} treats all layers identically during the gradient descent meta-update.} GLO allows us to prioritize learning shallow layers prior to deeper layers providing fine-grained (layer level) control over training dynamics in contrast to traditional IE approaches. Specifically, at each iteration \(r \in \{1,2,\dots,\mathcal{R}\}\), the (binary) gating value for layer \(l\) is calculated with a gating mechanism detailed in Eq.~\ref{eq:linear_gating}.
\begin{equation}
  g^l(r) =
  \begin{cases}
    1, & l \;\le\; \bigl\lfloor\tfrac{r\,L}{\mathcal{R}}\bigr\rfloor,\\
    0, & \text{otherwise},
  \end{cases}
  \quad l=0,\dots,L-1.
\label{eq:linear_gating}
\end{equation}
The gating schedule begins with the first layer unmasked, followed by linear unmasking of deeper layers as training progresses, ensuring localized (and more stable) effects of non-trivial loss dynamics.  
Although the proposed $g(r) \in \{0,1\}^L$, results in a linear gating schedule, GLO can easily admit any general gating mechanism.


\vspace{-1.5ex}
\begin{equation}
    \theta^{l}_{r+1} =\theta^{l}_r -\eta_{\mathrm{outer}}\,g^l(r)\,\nabla_{\theta_r^l}\mathcal{L}_{\mathrm{meta}},
    \quad l=0,\dots,L-1,
    \label{eq:glo_layer_wise_meta_update}
\end{equation}

If $g(r) = [g^0(r),\,\dots,\,g^{L-1}(r)]$ and each meta-update gradient vector is represented as $\nabla_{\Theta_r}\mathcal{L}_{\mathrm{meta}}
=[\nabla_{\theta^0_r}\mathcal{L}_{\mathrm{meta}},\,\dots,\,\nabla_{\theta_r^{L-1}}\mathcal{L}_{\mathrm{meta}}]$, then the meta-update at iteration $r+1$ for layer $l$ is detailed in Eq.~\ref{eq:glo_layer_wise_meta_update}.
Here, each \(g^l(r)\) controls the meta-update strength of layer \(l\).  
Alg.~\ref{alg:mlayer_iif} details the entire procedure of our proposed~\mymethod{} framework.\hide{for learned initialization of PINN weights.} The learned weights $\Theta_\mathcal{R}$ resulting from Alg.~\ref{alg:mlayer_iif},\hide{obtained from the IE + GLO based process} can be employed as initial weights for PINN training on $\gamma_{\mathrm{hard}}$ (i.e., the \emph{hard} PDE context).

%% file: sections/c4.ProblemFormulation/dataset.tex
\section{Dataset Description and Experimental Setup}\label{sec:dataset}
\vspace{-1.5ex}
We evaluate our proposed \mymethod{} model on three diverse PDE domains, described below.

\par\noindent
\textbf{1D Convection.}
The 1D convection PDE (Eq.~\ref{eq:conv}) models transport of a scalar field $u(z,t)$ with constant velocity. In Eq.~\ref{eq:conv}, $z\in[0,2\pi]$ denotes the spatial coordinate and  $t \in[0,1]$ denotes the temporal coordinate. $\beta$ is the advection coefficient, 
$u_t$ and $u_z$ denote partial derivatives in time and space. 
$u(z,0)=sin(z)$ defines the initial condition and periodic boundaries impose $u(0,t)=u(2\pi,t)$. 
\vspace{-0.5ex}
\begin{equation}
\mathcal{N}[u](z,t) := u_t(z,t) + \beta\,u_z(z,t) = 0,
\label{eq:conv}
\end{equation}

This problem represents convection-dominated motion, where high-speed phase propagation and increasingly oscillatory space-time structure make PINN training challenging.
In line with prior work~\cite{krishnapriyan_characterizing_2021}, we consider $\Gamma_{\mathrm{easy}}$ as $\beta\in\{5,10,15,20,25\}$ and hard regime as $\beta\in\{40,50,60,70,80\}$. Selection methodology for easy, hard tasks is detailed in Appendix~\ref{app:task_difficulty_boxplots}.

\par\noindent
\textbf{2D Helmholtz.}
The steady-state Helmholtz PDE (Eq.~\ref{eq:helm}) describes oscillatory wave fields on a 2D domain.
Let $z=(x_1,x_2)$ and let $\Delta=\partial_{x_1x_1}+\partial_{x_2x_2}$ denote the Laplace operator. Eq.~\ref{eq:helm} is defined on the domain $[-1,1]^2$ with Dirichlet boundary condition $u=0$ on $\partial\Omega$.
Using the analytic solution $u(z)=\sin(a_1\pi x_1)\sin(a_2\pi x_2)$, the source term becomes
$q(z)=\big[k^2-(a_1\pi)^2-(a_2\pi)^2\big]\,u(z)$.
\begin{equation}
\mathcal{N}[u](z) := \Delta u(z) + k^2 u(z) - q(z) = 0,
\label{eq:helm}
\end{equation}
The coefficients $(a_1,a_2)$ control the oscillation frequency along x and y direction respectively, and larger values induce more frequent oscillations posing greater challenge for PINNs. $\Gamma_{\mathrm{easy}}$ is considered $\{(1,1),(1,2),(1,3),(1,4),(1,5)\}$ based on the case considered in~\cite{wang2021understanding} and hard regime is selected to have significantly more challenging propagation dynamics and both isotropic and anisotropic propagation dynamics are tested in the hard regimes
$\{(4,4),(4,5),(5,5),(4,6),(5,6),(6,6)\}$. Our work has significantly expanded the investigation regime, compared to previous PINN-based investigations of 2D Helmholtz. Selection methodology for easy, hard tasks is detailed in Appendix~\ref{app:task_difficulty_boxplots}..

\par\noindent
\textbf{2D Incompressible Navier--Stokes.}
We finally consider modeling the challenging 2D viscous flow over a cylinder, using the non-dimensional 2D incompressible Navier--Stokes equations, represented in Eq.~\ref{eq:ns} in compact momentum-continuity form. Here $\mathbf{u}(x,y,t)$ is the velocity field, $p(x,y,t)$ is the scalar pressure field,
$\nu$ is the kinematic viscosity (parameterized through the Reynolds number). Domain $(x,y,t)\in\Omega\times[0,T]$, where $\Omega$ contains a unit-diameter cylinder.
\begin{equation}
\frac{\partial \mathbf{u}}{\partial t} + (\mathbf{u}\cdot\nabla)\mathbf{u} + \nabla p  - \nu \nabla^{2}\mathbf{u}
= 0,
\qquad
\nabla\cdot\mathbf{u}=0.
\vspace{-0.3ex}
\label{eq:ns}
\end{equation}
This system exhibits vortex shedding and transitional flow behavior, and the coupling between
momentum and incompressibility creates multiple interacting residual terms, making it one of the most challenging PDE families for PINNs. The complexity of the dynamics increases with increase in Reynolds number ($Re$).
We consider $\Gamma_{\mathrm{easy}}$ to be $\mathrm{Re}\in\{100,200,300,400,500\}$ where relatively more steady, laminar dynamics are exhibited and in line with~\citep{lee2025physics}, we consider 
$\mathrm{Re}\in\{600,800,1000\}$ to be hard regime where less steady dynamics are exhibited.

To ensure a \textbf{fair evaluation}, all PINN baseline methods are trained for 56K epochs on the target domain $\Gamma_{\mathrm{hard}}$, while \mymethod{} is trained only for 50K epochs on the target domain. This difference accounts for the 6K iterations employed for the learned initialization procedure which is absent in baseline methods which are randomly initialized.
\hide{are trained for the \textbf{same number of epochs (50k)}.} All other training parameters (detailed in Appendix~\ref{appendix:experimental-setup} for full reproducibility) are identical including model architecture, training data, optimization settings.\hide{This includes both standard PINN baselines and methods trained in a single-stage optimization setting.} \hide{Additionally, even when trained on a larger number of epochs (i.e., > 50K), we find that PINNs that undergo convergence failures do not improve their performance. These additional results are reported in Appendix~\ref{appen:extended-training}.} 

%% file: sections/c6.ResultsAnalysis/results_and_discussion.tex
\section{Results \& Discussion}\label{sec:results}
\vspace{-1.5ex}
In this section, we demonstrate with rigorous quantitative and qualitative experiments, the effects of catastrophic failures on PINN performance and how these failures can be alleviated via Learned Initialization.
To validate our approach, we compare \mymethod{} against a suite of state-of-the-art models, that have been designed specifically to overcome catastrophic failures of PINNs in challenging PDE domains; the full descriptions of these baselines can be found in Appendix~\ref{appendix_hessian_helm}.
Specifically, we undertake this investigation by answering three research questions (RQ):
\par\noindent
\textbf{RQ1.} (Qualitative) How does learned initialization alleviate catastrophic PINN training dynamics?
\par\noindent
\textbf{RQ2.} (Quantitative) How does \mymethod{} compare with other state-of-the-art (SoTA) approaches proposed to alleviate PINN catastrophic failures?
\par\noindent
\textbf{RQ3.} (Ablation) What is the effect of invariance encoding and gated layer-wise optimization in~\mymethod{}?
\hide{We evaluate our approach on three representative PDE systems: the 1D convection equation, the 2D Helmholtz equation, and the 2D time‐dependent incompressible Navier–Stokes equations. Full specifications of each setup, including PDE definitions, boundary and initial conditions, PDE Parameters, and training configurations, are provided in Section~\ref{sec:dataset}.}\hide{To ensure reproducibility we also detail training information such as model hyperparameters, architecture details in Appendix.~\ref{appendix:experimental-setup}.}
\vspace{-0.5ex}
\input{sections/c6.ResultsAnalysis/r1}

\vspace{-0.5ex}
\input{sections/c6.ResultsAnalysis/r2}

\vspace{-0.5ex}
\input{sections/c6.ResultsAnalysis/r3}
\vspace{-0.5ex}
\input{sections/c6.ResultsAnalysis/additional}

%% file: sections/c6.ResultsAnalysis/r1.tex
\subsection{(RQ1) How does learned initialization alleviate catastrophic PINN training dynamics?}

We answer this question in three stages. \ul{First}, we demonstrate qualitatively the catastrophic failures PINNs undergo in challenging PDE contexts. \ul{Second}, we investigate how the training losses of the PINNs evolve in challenging domains. \ul{Third}, we characterize the effect of the catastrophic training dynamics by analyzing the loss landscape neighborhood of the converged models. In each case, we contrast the analysis of a randomly initialized PINN, with \mymethod{} behavior in the same context.
\begin{figure}[!ht]
  \centering
  \begin{subfigure}[t]{0.44\textwidth}
    \centering
    \includegraphics[width=\textwidth]{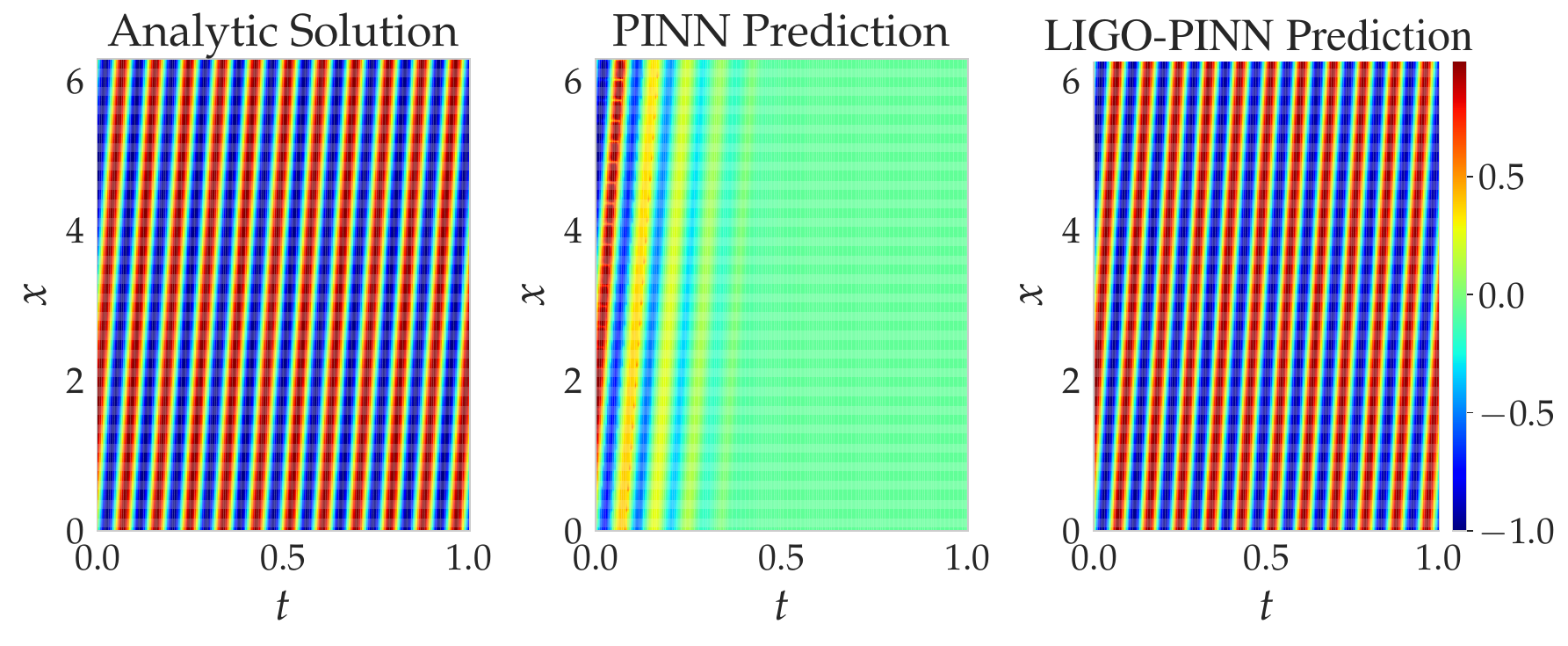}
    \caption{1D Convection ($\beta$=70)}
    \label{fig:heat_pinn_convergence}
  \end{subfigure}
  \hfill
  \begin{subfigure}[t]{0.44\textwidth}
    \centering
    \includegraphics[width=\textwidth]{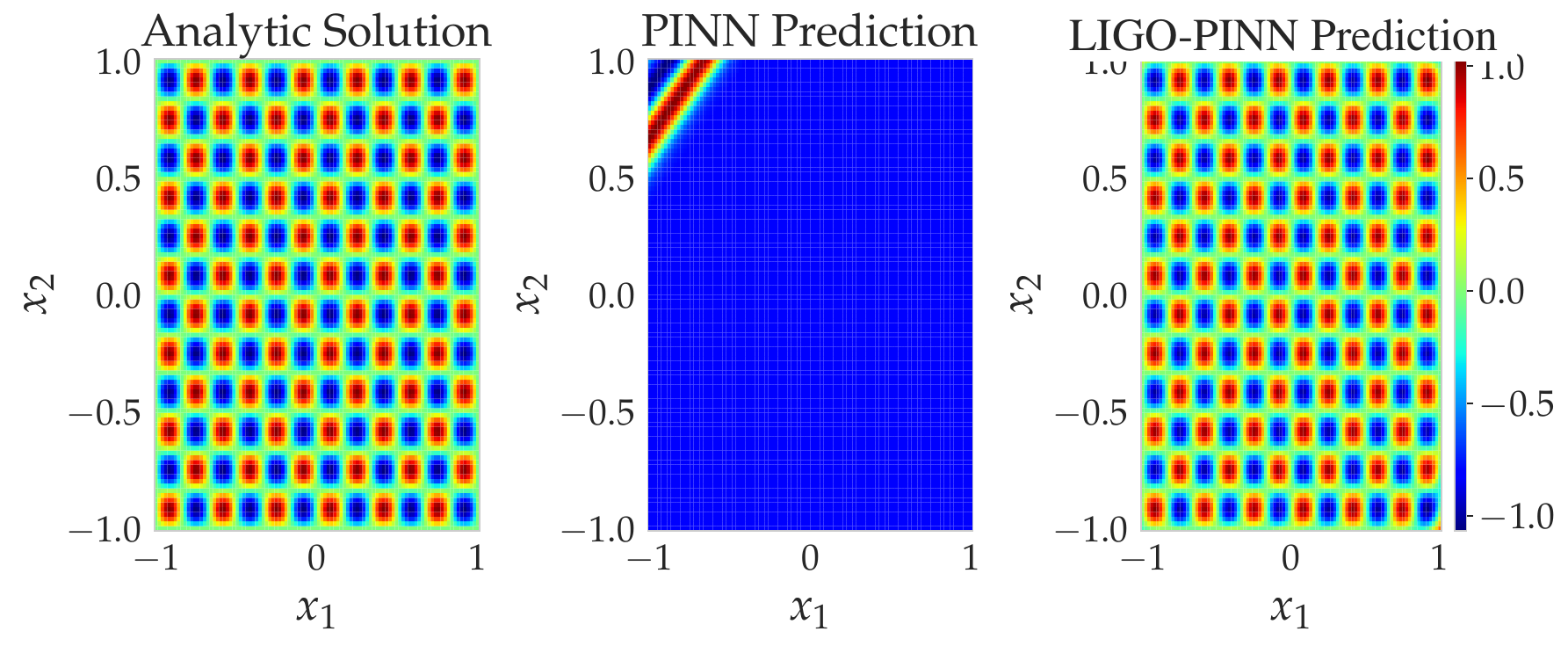}
    \caption{2D Helmholtz ($a_1 =\{6\}, a_2 = \{6\}$)}
    \label{fig:helm_pinn_convergence}
  \end{subfigure}
  \caption[PINN Convergence Heatmaps]{Qualitative performance comparison of a randomly initialized PINN model and \mymethod{} on two different and challenging PDE domains. We notice that both in the 1D convection (Fig.~\ref{fig:pinn_convergence_heatmaps}a) and 2D Helmholtz (Fig.~\ref{fig:pinn_convergence_heatmaps}b), \mymethod{} faithfully recreates the analytical solution while the randomly initialized PINN variant experiences catastrophic failure.}
  \label{fig:pinn_convergence_heatmaps}
  \vspace{-3ex}
\end{figure}

\textbf{Qualitative Failure Modes.}
In Fig.~\ref{fig:pinn_convergence_heatmaps}a, we demonstrate the solution predicted by a PINN (with random weight initialization) when trained on a challenging 1D convection domain with $\beta = 70$. It can be clearly seen that the PINN model (center plot in Fig.~\ref{fig:pinn_convergence_heatmaps}a) has failed to estimate a good solution of the domain. The perfect solution of the domain can be seen in the left plot (labeled `Analytical Solution') of Fig.~\ref{fig:pinn_convergence_heatmaps}a.  We conduct a similar qualitative investigation of a separate PDE domain 2D-Helmholtz comprising two PDE parameters $a_1, a_2$, with $(a_1 = 6, a_2 = 6)$ in Fig.~\ref{fig:pinn_convergence_heatmaps}b. Once again, we notice that the PINN model (center plot of Fig.~\ref{fig:pinn_convergence_heatmaps}b) is unable to provide a good estimate of the challenging PDE domain solution (i.e., `Analytical Solution' represented on the left plot in  Fig.~\ref{fig:pinn_convergence_heatmaps}b). Turning our attention to the performance of~\mymethod{} in each of these two contexts (i.e., right plot in each of Fig.~\ref{fig:pinn_convergence_heatmaps}a and Fig.~\ref{fig:pinn_convergence_heatmaps}b), we notice that the estimation of~\mymethod{} closely aligns with the analytical solution of the corresponding PDE domain. We now highlight that the only difference between the PINN and \mymethod{} procedures is the weight initialization. Specifically, PINN weights have been initialized randomly (i.e., Xavier initialization) while the weights of \mymethod{} have been initialized by our proposed \emph{learned initialization} method.\hide{Once weights have been initialized, both models have been subjected to identical training procedures on the target PDE domain.} Thus, we infer that the learned initialization strategy in \mymethod{} successfully alleviates catastrophic failures to which traditional PINNs are susceptible, when modeling complex PDE dynamics.
\vspace{-1ex}
\par\noindent
\ul{Extension to Multi-Physics.}
We also investigate \mymethod{} performance on 2D incompressible, Navier-Stokes cylinder wake at high Reynolds numbers. A challenging multi-physics setting. 
We investigate both forward (pressure prediction) and inverse settings (recovering viscous and advection coefficients of Navier-Stokes equations).\hide{velocity component prediction $(u,v)$, while pressure is predicted in a physics-informed manner.}\hide{In the inverse setting, the advection and viscosity coefficients $(\lambda_{\text{adv}}, \lambda_{\text{vis}})$ are treated as learnable parameters.} 
\hide{Domain visualizations and pressure prediction results are provided in Appendix~\ref{appendix:ns_details}.}
\begin{figure}[!htbp]
\vspace{-1.3ex}
  \centering
  \makebox[\textwidth][c]{%
    \begin{subfigure}[b]{0.30\textwidth}
      \centering
      \includegraphics[width=\textwidth]{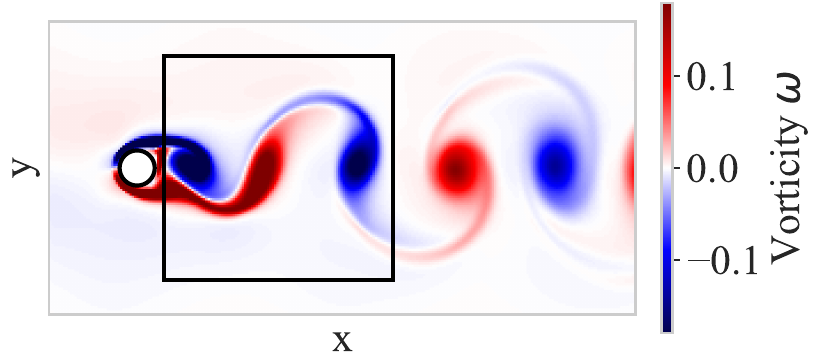}
      \caption{Flow past a cylinder with vortex shedding (FEM solver)}
      \label{fig:ns_full_slice}
    \end{subfigure}\hspace{1em}%
    \begin{subfigure}[b]{0.30\textwidth}
      \centering
      \includegraphics[width=\textwidth]{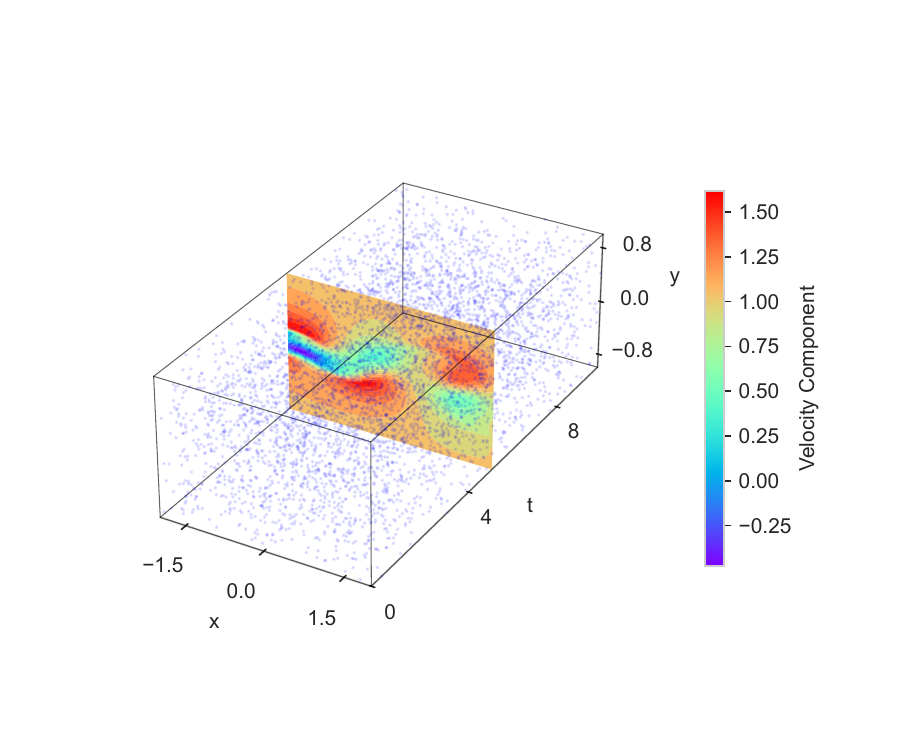}
      \caption{Streamwise velocity $u_x$ at $t$.}
      \label{fig:ns_vorticity_slice}
    \end{subfigure}\hspace{1em}%
    \begin{subfigure}[b]{0.30\textwidth}
      \centering
      \includegraphics[width=\textwidth]{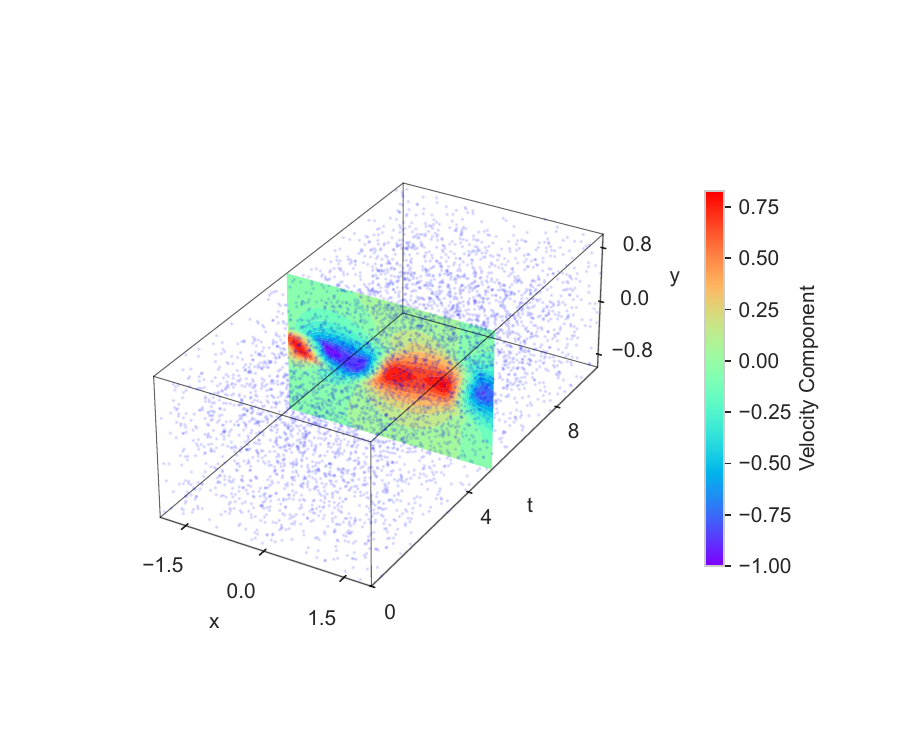}
      \caption{Transverse velocity $u_y$ at $t$.}
      \label{fig:ns_pressure_slice}
    \end{subfigure}%
  }

  \vspace{1em} 

  \makebox[\textwidth][c]{%
    \begin{subfigure}[t]{0.30\textwidth}
      \centering
      \includegraphics[width=0.60\textwidth]{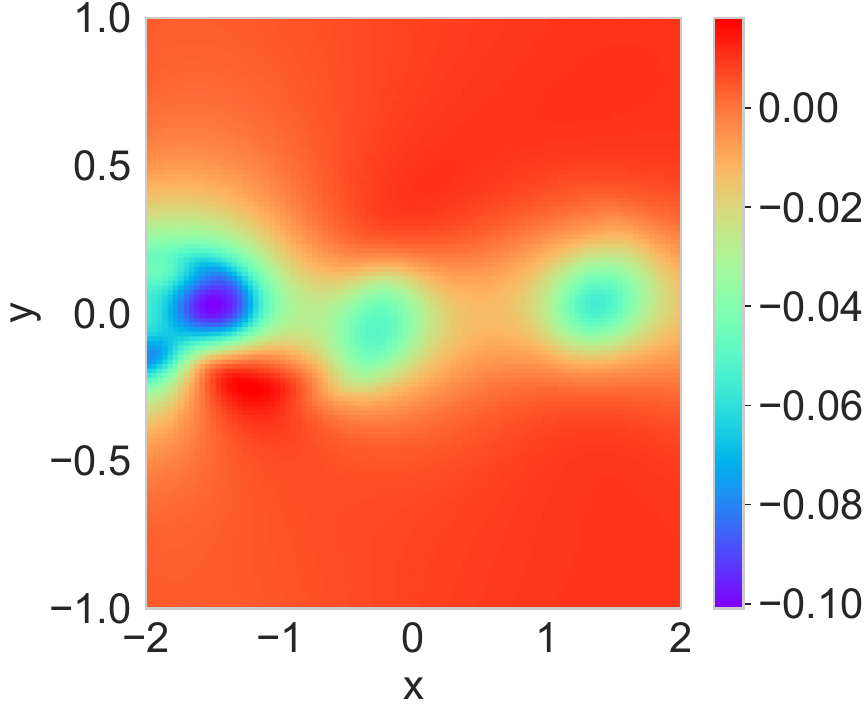}
      \caption{Ground-truth pressure $p$.}
      \label{fig:ns_pressure_gt}
    \end{subfigure}\hspace{1em}%
    \begin{subfigure}[t]{0.30\textwidth}
      \centering
      \includegraphics[width=0.60\textwidth]{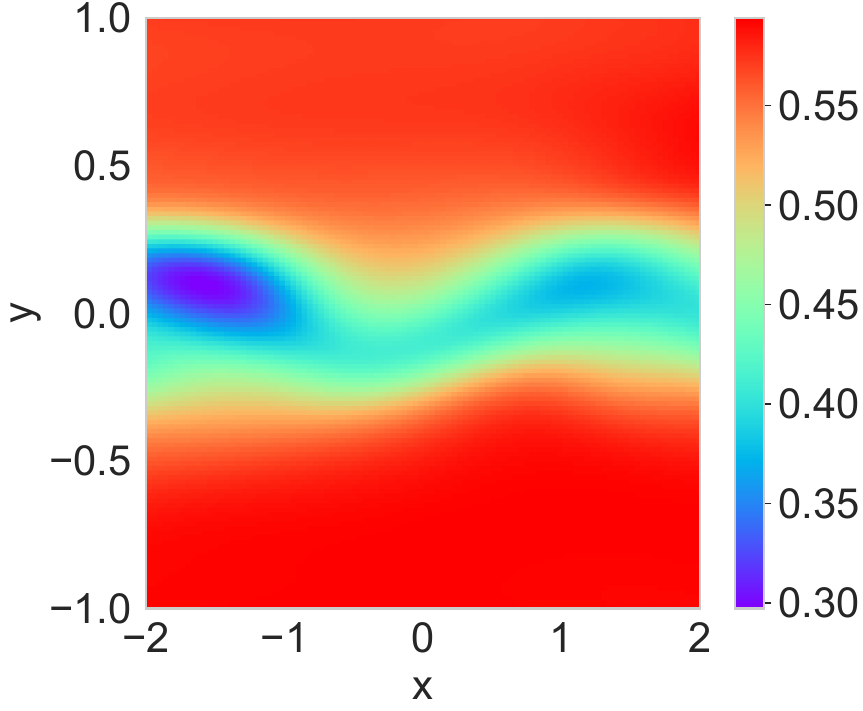}
      \caption{PINN pressure (failure).}
      \label{fig:ns_pinn_error}
    \end{subfigure}\hspace{1em}%
    \begin{subfigure}[t]{0.30\textwidth}
      \centering
      \includegraphics[width=0.60\textwidth]{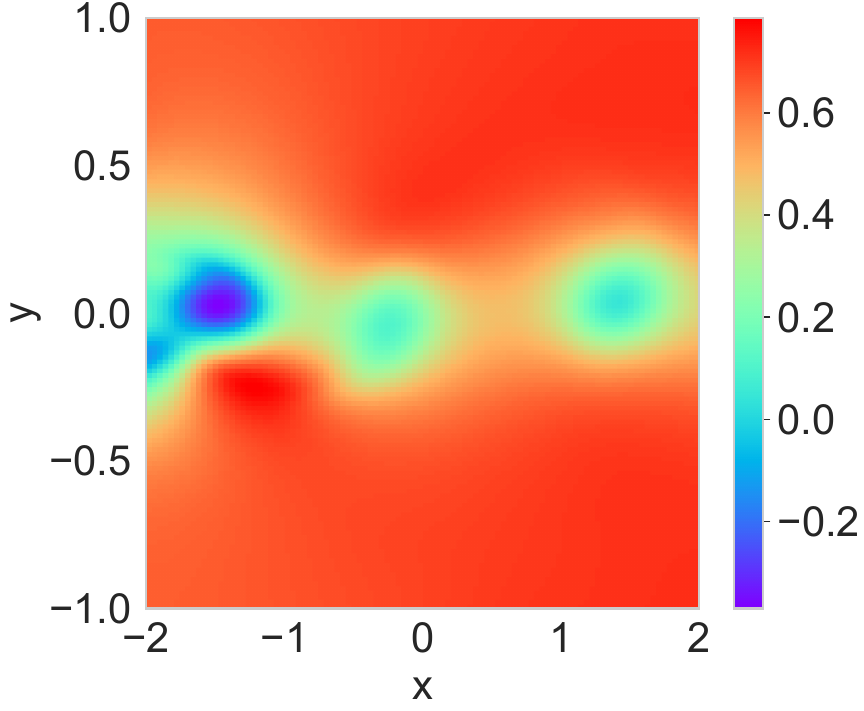}
      \caption{LIGO-PINN pressure.}
      \label{fig:ns_LIGOpinn_error}
    \end{subfigure}%
  }

  \caption[Navier-Stokes training and evaluation at $\mathrm{Re}=1000$]{ 
Navier-Stokes cylinder wake at $\mathrm{Re}=1000$.  
(a) Simulation setup with vortex shedding (FEM solver). The black box indicates the region used for training and evaluation.  
(b--c) Velocity components $u_x, u_y$ at a representative time slice. Also showcased are collocation points at which velocity training data is sampled.
(d) Ground-truth pressure.  
(e) Standard PINN prediction (failure).  
(f) LIGO-PINN prediction, showing faithful reconstruction of ground-truth pressure.
\vspace{-3ex}
}

  \label{fig:ns_convergence_slices}
\end{figure}

\hide{We evaluate the 2D incompressible, time-dependent Navier--Stokes system at $\mathrm{Re}=1000$.} 
\hide{In Fig.~\ref{fig:ns_convergence_slices}, we clearly see that vanilla PINNs fail to reproduce the pressure field Fig.~\ref{fig:ns_convergence_slices}(e), while \mymethod{} reconstructs the wake structures faithfully Fig.~\ref{fig:ns_convergence_slices}(f).}  
Consistent with our findings on 1D-Convection and 2D-Helmholtz, we see from Fig.~\ref{fig:ns_convergence_slices}, that \mymethod{} yields substantially lower errors and qualitatively faithful pressure fields (Fig.~\ref{fig:ns_LIGOpinn_error}) compared to randomly initialized PINN (Fig.~\ref{fig:ns_pinn_error}). Ground truth pressure field is depicted in Fig.~\ref{fig:ns_pressure_gt}.
This highlights the advantage of learned initialization to model complex fluid dynamics.
\begin{figure}[!ht]
  \centering
  \begin{subfigure}[t]{0.3\textwidth}
    \centering
    \includegraphics[width=0.7\textwidth]{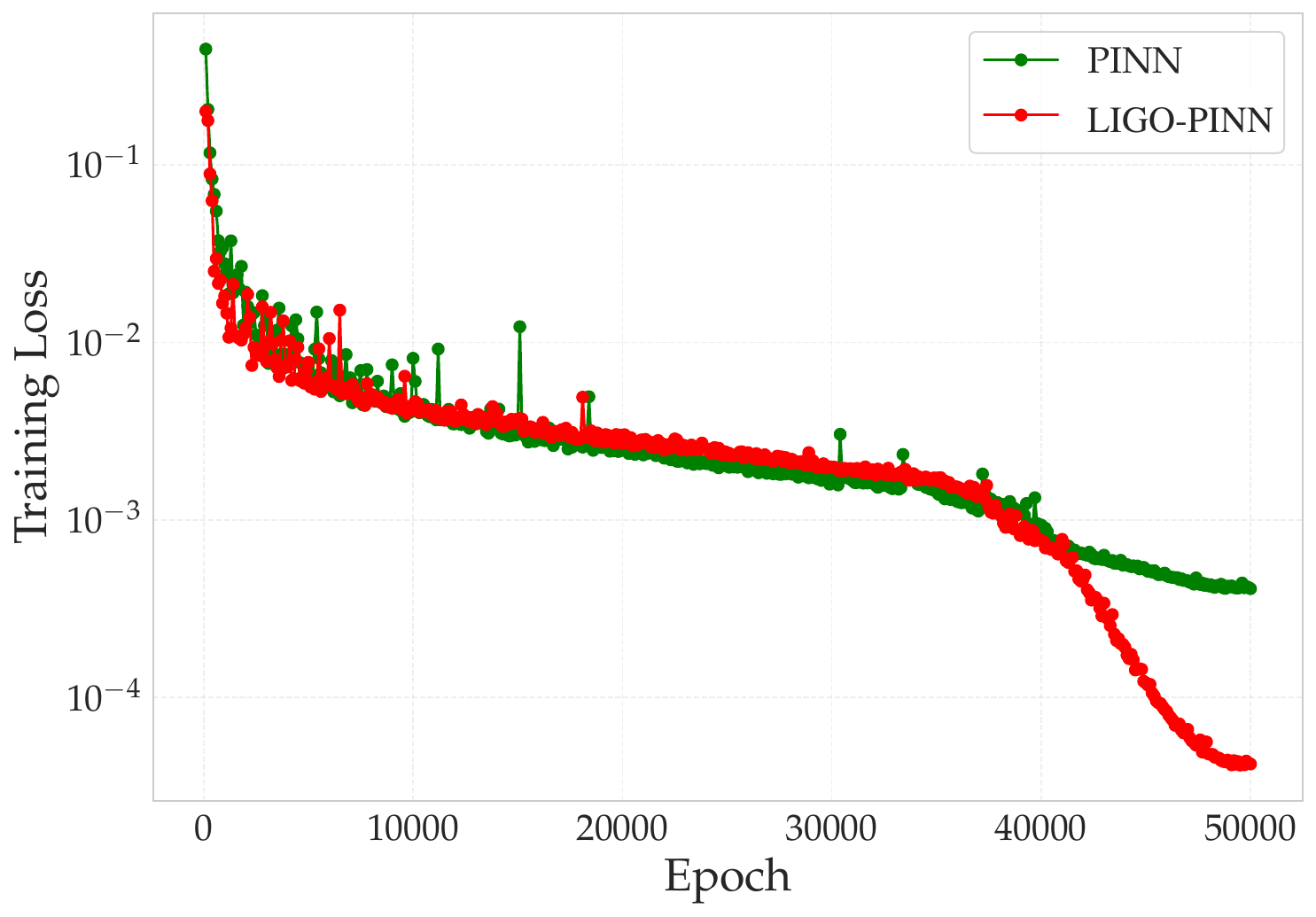}
    \caption{1D Convection ($\beta = 70$)}
    \label{fig:heat_pinn_dynamic}
  \end{subfigure}
  \hfill
  \begin{subfigure}[t]{0.3\textwidth}
    \centering
    \includegraphics[width=0.7\textwidth]{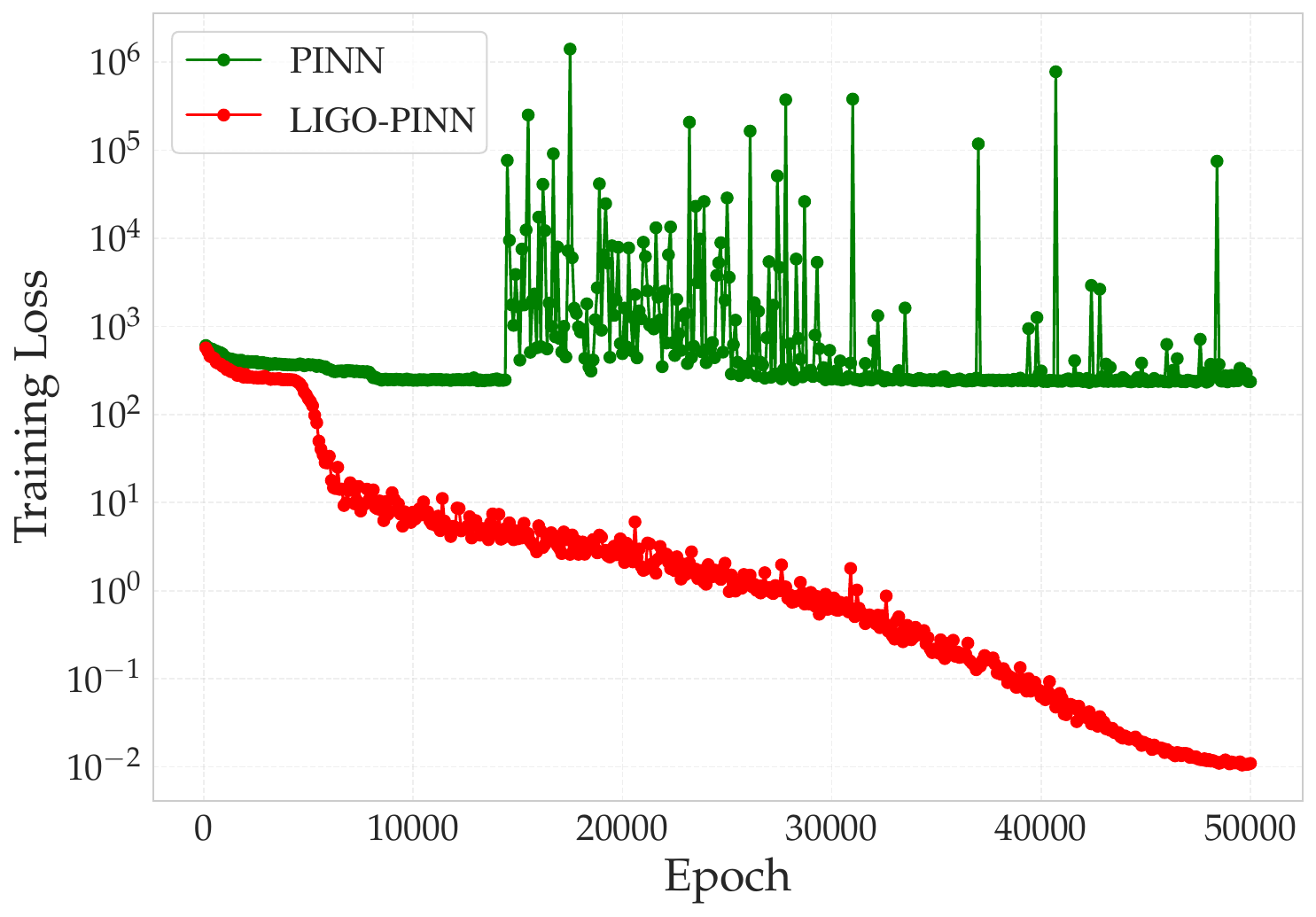}
    \caption{2D Helmholtz ($a_1$= 6, $a_2$= 6)}
    \label{fig:helm_pinn_dynamic}
  \end{subfigure}
  \hfill
  \begin{subfigure}[t]{0.3\textwidth}
    \centering
    \includegraphics[width=0.7\textwidth]{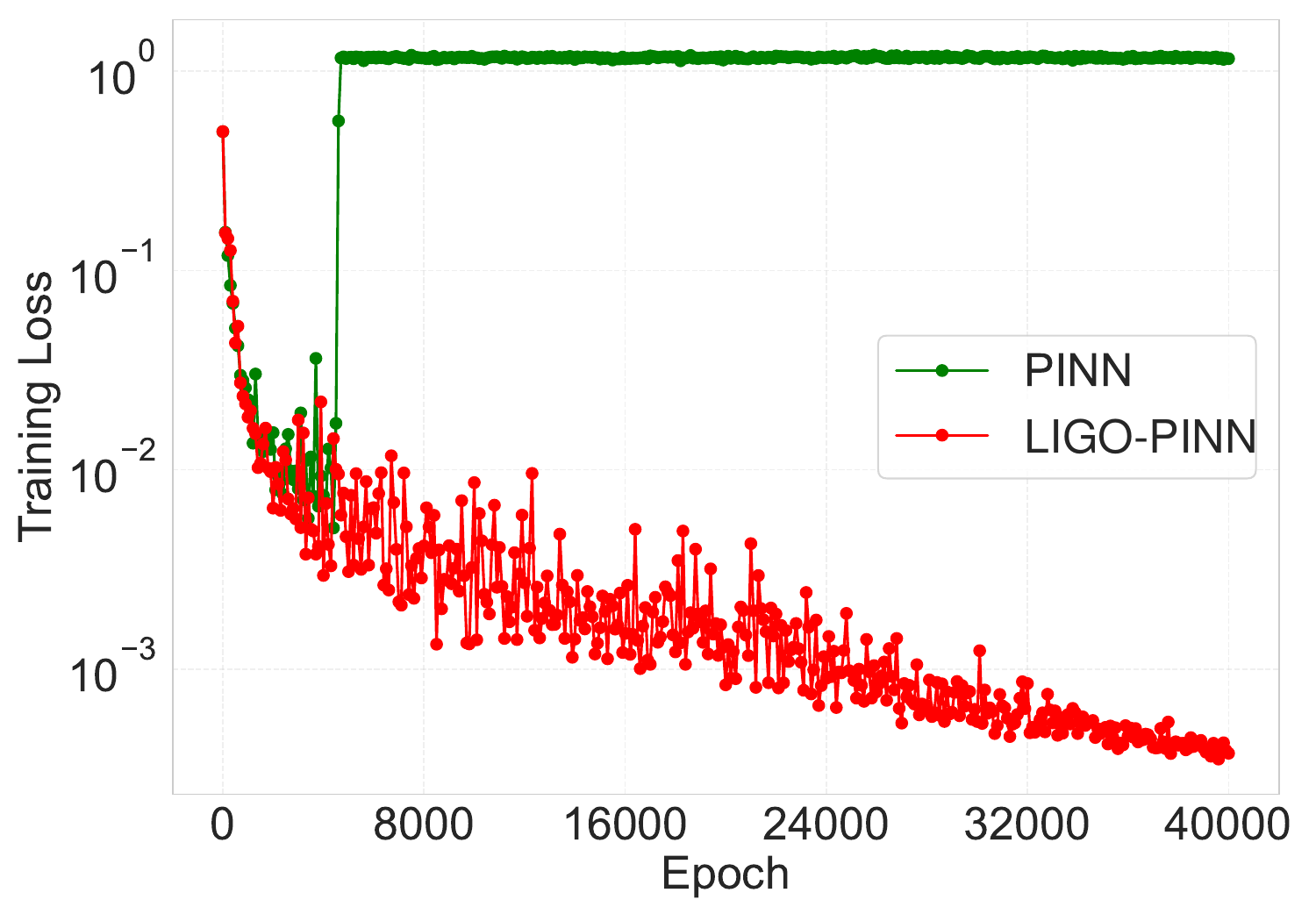}
    \caption{2D Navier-Stokes ($Re$= 1000)}
    \label{fig:ns_pinn_dynamic}
  \end{subfigure}
  
  \caption[PINN dynamic losses]{Loss curves highlighting training dynamics of randomly initialized PINNs and ~\mymethod{}, across three PDE domains. In all three cases, ~\mymethod{} converges to a solution with (at least) an order of magnitude lower loss, compared to randomly initialized PINN.}
  \label{fig:pinn_dynamic_losses}
  \vspace{-4ex}
\end{figure}

\par\noindent
\textbf{Training Dynamics Analysis.}
We complement the qualitative analysis  in Fig.~\ref{fig:pinn_convergence_heatmaps} with an investigation (Fig.\ref{fig:pinn_dynamic_losses}) of the training dynamics in the same PDE contexts. Specifically, we highlight in Fig.\ref{fig:pinn_dynamic_losses}, the evolution of the total training loss (Eq.~\ref{eq:pinn-total-loss}) of PINNs and \mymethod{} on the same 1D convection, 2D Helmholtz and 2D Navier-Stokes domains.
We note that the training loss plots in Fig.\ref{fig:pinn_dynamic_losses} has been averaged over five independent training runs to capture the \emph{expected training dynamics}.
\hide{\textcolor{red}{Loss plots of individual training runs are  in Appendix~\ref{appendix:extended_results}}.} 

For the 1D convection problem  ($\beta = 70$) the result depicted in  Fig.~\ref{fig:pinn_dynamic_losses}a, shows that random initialization of the PINN leads to an early plateau in the loss curve, whereas our learned initialization method (\mymethod{}) achieves a steady, monotonic decrease, signaling more effective exploration of the solution manifold and ultimately, convergence to a loss value an order of magnitude lower than the corresponding randomly initialized PINN solution. A similar pattern emerges on the 2D Helmholtz and 2D Navier-Stokes problem $(a_1=6,\,a_2=6)$ as shown in Fig.~\ref{fig:pinn_dynamic_losses}b,c. We see that the randomly initialized PINN exhibits premature convergence to a sub-optimal solution while, once again \mymethod{} achieves a loss \emph{several orders of magnitude lower}. This  demonstrates the ability of our learned initialization based solution (\mymethod{}) to achieve superior convergence dynamics by overcoming loss surface barriers~\cite{fort2020deep} and hence avoiding catastrophic failures.
\begin{figure}[!htbp]
  \centering
  \begin{subfigure}[t]{0.7\textwidth}
    \centering
    \includegraphics[width=\textwidth]{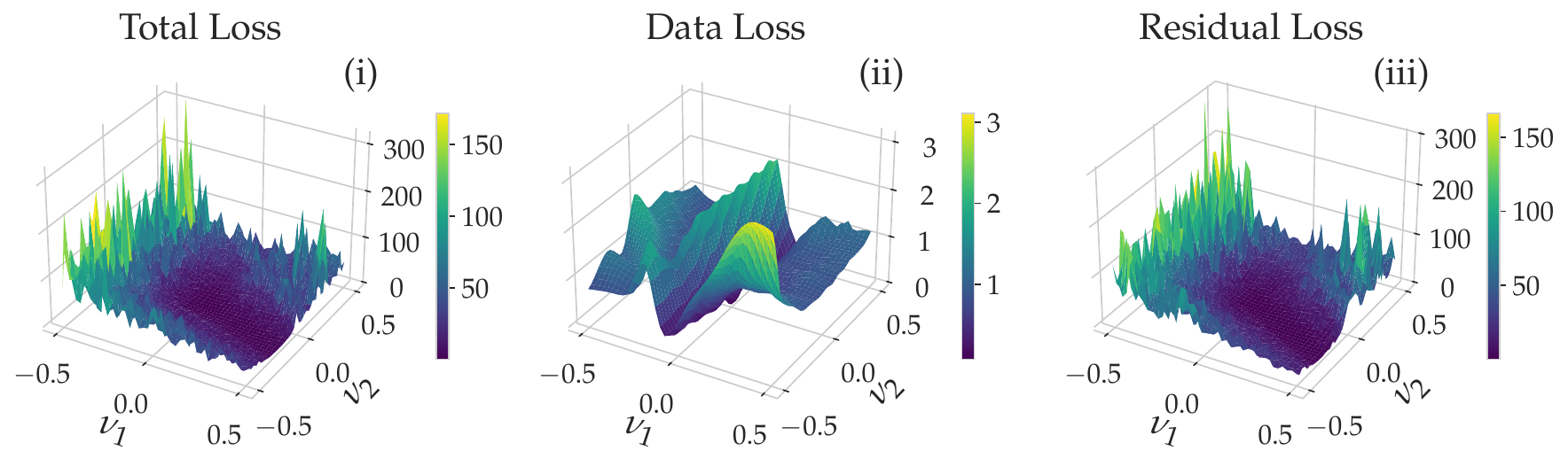}
    \caption{1D Convection PINN Loss Landscape}
    \label{fig:heat1d_surface}
  \end{subfigure}
  \vfill
  \begin{subfigure}[t]{0.7\textwidth}
    \centering
    \includegraphics[width=\textwidth]{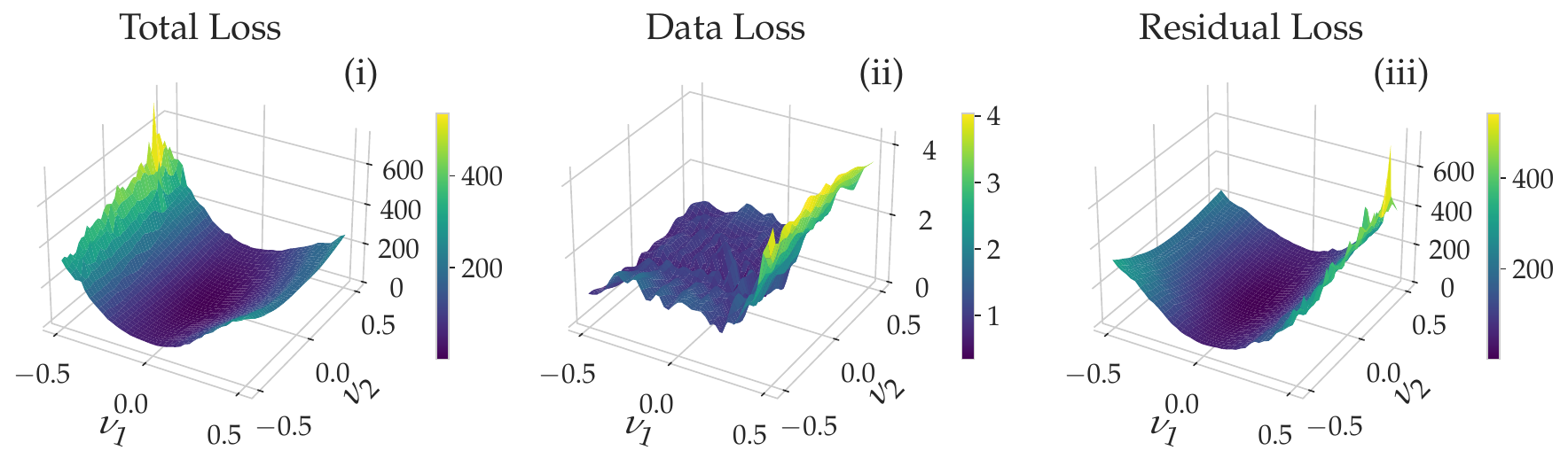}
    \caption{1D Convection \mymethod{} Loss Landscape}
    \label{fig:helm2d_surface}
  \end{subfigure}
  \caption[PINN Spectral Surfaces]{%
    Loss Landscape visualization generated by perturbing converged model weights along its top-2 eigenvectors ($\nu_{1}$, $\nu_{2}$). 
    (a) Randomly Initialized PINN 1D Convection, with (i), (ii), (iii) representing loss landscape w.r.t total loss, data loss and residual loss respectively. (b) \mymethod{} 1D Convection with (i), (ii), (iii) once again representing loss landscape w.r.t total, data, residual loss. }
  \label{fig:pinn_loss_landscape}
\vspace{-2ex}
\end{figure}

\textbf{Loss Landscape Analysis.}
Although the loss curves in Fig.~\ref{fig:pinn_dynamic_losses} show a clear contrast between random and learned initialization, they do not fully reveal how training dynamics affect parameter optimization. To clarify this, we examine the loss landscape neighborhood of each \emph{converged} model using a Hessian‐informed perturbation analysis~\citep{krishnapriyan_characterizing_2021,B_ttcher_2024}. Fig.~\ref{fig:pinn_loss_landscape} presents results for the 1D convection problem (2D Helmholtz in Appendix~\ref{appendix_hessian_helm}). Under random initialization (Fig.~\ref{fig:pinn_loss_landscape}a), the converged PINN exhibits a highly rugged surface with deep valleys and sharp ridges, driven mainly by the residual term (Fig.~\ref{fig:pinn_loss_landscape}a.iii). The presence of such ravines is known to hinder convergence to strong local minima~\cite{wu2017towards,hochreiter1997flat}. In contrast, the data loss landscape is smoother but still challenging to optimize.  With~\mymethod{}, the converged landscape is markedly smoother without the sharp peaks of Fig.~\ref{fig:pinn_loss_landscape}a. Fig.~\ref{fig:pinn_loss_landscape}b.i shows the total loss surface, while Figs.~\ref{fig:pinn_loss_landscape}b.ii--b.iii show data and residual components, all with smoother geometry compared to their counterparts resulting from a randomly initialized PINN. Viewed together with Fig.~\ref{fig:pinn_dynamic_losses}a, these results illustrate that random initialization produces residual-dominated peaks that trap the optimizer in narrow basins, whereas~\mymethod{} yields a smoother landscape that supports improved convergence to better local minima.

%% file: sections/c6.ResultsAnalysis/r2.tex
\subsection{(RQ2) How do LIGO-PINNs compare with other state-of-the-art (SoTA) approaches proposed to alleviate PINN catastrophic failures?}
Recent efforts have begun to investigate catastrophic failures in PINNs, and in this section we compare our learned-initialization-based~\mymethod{} model against six comparative PINN baselines, including vanilla PINNs and recent methods proposed to improve PINN training stability~\citep{krishnapriyan_characterizing_2021,wang2021understanding,daw_mitigating_2023,qin2022metapdelearningsolvepdes,cho2023hypernetwork}. 
We provide quantitative comparisons using the mean-absolute error (MAE) metric, across three benchmark PDE systems: 
the 1D Convection equation, 
the 2D Helmholtz equation, 
and the 2D incompressible time-dependent Navier-Stokes system~\citep{raissi2017physicspart1}. 
In all cases, we evaluate models under challenging PDE settings, and extend our analysis regime far beyond what existing papers have investigated.\hide{where PDE parameters or regimes lie outside those directly observed during initialization. }
For Navier-Stokes, we report both forward pressure reconstruction and inverse coefficient estimation in the main text.

\input{sections/c6.ResultsAnalysis/heat_table}
\textbf{1D Convection System}  
Table~\ref{tab:convection_mae_extrapolation} shows that~\mymethod{} achieves the lowest average MAE across the extrapolation settings and obtains the best result in four out of five tested regimes. 
~\mymethod{} achieves the strongest average performance, reducing average MAE by \textbf{96.5\%} relative to vanilla PINNs and by \textbf{68.9\%} relative to the strongest non-ours baseline, PINN-Dynamic. Across all six comparative baselines, the MAE reduction ranges from \textbf{68.9\% to 96.5\%}.  
\hide{We need that \mymethod{} seed 50k training epoch the evaluation task compared to 56k that was seen by the PINN baseline}




\input{sections/c6.ResultsAnalysis/helm_table}
\vspace{-2ex}
\textbf{2D Helmholtz System.}  
The 2D Helmholtz system presents a more challenging oscillatory and multi-parameter setting. As shown in Table~\ref{tab:helmholtz_mae_methods_flat}, the baseline methods remain substantially less accurate than~\mymethod{} across all tested domain parameters. On average,~\mymethod{} reduces MAE by \textbf{94.3\%} relative to vanilla PINNs and by \textbf{92.6\%} relative to the strongest non-ours baseline, PINN-Dynamic. Across all six comparative baselines, the MAE reduction ranges from \textbf{92.6\% to 95.5\%}. These results indicate that learned initialization is particularly effective for high-frequency PDE regimes where standard PINN training and existing mitigation strategies remain unstable.
\vspace{-0.5ex}

\input{sections/c6.ResultsAnalysis/ns_table}

\vspace{-1ex}
\textbf{2D Incompressible Navier--Stokes.}  Table~\ref{tab:ns_forward} reports the pressure prediction results across Reynolds numbers. In this fluid dynamics setting,~\mymethod{} achieves an average MAE of $0.0096$, compared with $0.0517$ of \emph{PINN (dynamic)}, (the strongest baseline), and $0.0791$ for vanilla PINNs (i.e., PINN (fixed)). This corresponds to an error reduction of \textbf{81.4\%} relative to the strongest baseline and \textbf{87.9\%} relative to vanilla PINNs. Across all six comparative baselines, the MAE reduction ranges from \textbf{81.4\% to 97.4\%}. These results show that learned initialization remains effective even in complex time-dependent fluid dynamics problems.

In addition to the forward pressure reconstruction setting, we further evaluate all methods on the inverse formulation of the 2D incompressible Navier--Stokes system.
In this setting, the advection and viscosity coefficients $(\lambda_{\mathrm{adv}}, \lambda_{\mathrm{visc}})$ are treated as unknown parameters and jointly inferred during training from sparse observations.
This inverse problem is particularly challenging due to the strong coupling between nonlinear momentum dynamics and incompressibility constraints, especially at higher Reynolds numbers.
Table~\ref{tab:ns_inverse} reports the coefficient estimation errors across Reynolds numbers $\mathrm{Re}\in\{600,800,1000\}$.
We observe that standard PINN baselines exhibit unstable parameter recovery and large estimation errors as the flow dynamics become increasingly complex.
In contrast,~\mymethod{} consistently achieves lower coefficient estimation error across all evaluated regimes, demonstrating that learned initialization improves not only forward field reconstruction but also inverse parameter identification in challenging multi-physics PDE systems.
For the inverse problem involving estimation of the advection ($\lambda_{\mathrm{adv}}$) and viscosity ($\lambda_{\mathrm{visc}}$) coefficients,
\textit{\mymethod{} yields an \textbf{80.2\%}, and \textbf{45.9\%} performance improvement over the strongest baseline (HyperLR in the inverse modeling case).}

\begin{table}[t]
\small
\setlength{\tabcolsep}{3pt}
\caption{Navier--Stokes: $\lambda_{\mathrm{adv}}$ and $\lambda_{\mathrm{visc}}$ estimation errors across Reynolds numbers. Lowest values in each column are in bold.}
\label{tab:ns_inverse}
\centering
\resizebox{\textwidth}{!}{%
\begin{tabular}{l*{3}{cc}|cc}
\toprule
\textbf{Method}
 & \multicolumn{2}{c}{\textbf{Re\_600}}
 & \multicolumn{2}{c}{\textbf{Re\_800}}
 & \multicolumn{2}{c}{\textbf{Re\_1000}}
 & \multicolumn{2}{c}{\textbf{Avg}} \\
\cmidrule(lr){2-3} \cmidrule(lr){4-5} \cmidrule(lr){6-7} \cmidrule(lr){8-9}
 & $\lambda_{\mathrm{adv}}$ & $\lambda_{\mathrm{visc}}$
 & $\lambda_{\mathrm{adv}}$ & $\lambda_{\mathrm{visc}}$
 & $\lambda_{\mathrm{adv}}$ & $\lambda_{\mathrm{visc}}$
 & $\lambda_{\mathrm{adv}}$ & $\lambda_{\mathrm{visc}}$ \\
\midrule

\methodpinnf
 & 0.1212 & 0.0282
 & 0.1210 & 0.0441
 & 0.0191 & 0.0074
 & 0.0871 & 0.0266 \\

\methodpinnd~\citep{wang2021understanding}
 & 0.0472 & 0.0085
 & 0.1539 & 0.0370
 & 0.0478 & 0.0074
 & 0.0830 & 0.0176 \\

\methodcurriculumn~\citep{krishnapriyan_characterizing_2021}
 & 0.6251 & 0.1645
 & 0.7021 & 0.1479
 & 0.6380 & 0.1787
 & 0.6551 & 0.1637 \\

\methodevo~\citep{daw_mitigating_2023}
 & 0.3841 & 0.0692
 & 0.5106 & 0.0722
 & 0.2222 & 0.1189
 & 0.3723 & 0.0868 \\

\methodmetapde~\citep{qin2022metapdelearningsolvepdes}
 & 0.4320 & 0.1312
 & 0.6910 & 0.2979
 & 0.6910 & 0.2370
 & 0.6046 & 0.2220 \\

\methodhyperlr
 & 0.0509 & 0.0108
 & 0.0633 & 0.0133
 & 0.1162 & 0.0163
 & 0.0768 & 0.0135 \\

 \methodmaml
 & 0.2679 & 0.0510
 & 0.2904 & 0.0374
 & 0.0519 & 0.0087
 & 0.2034 & 0.0324 \\

\textbf{\mymethod{} (Ours)}
 & \textbf{0.0291} & \textbf{0.0083}
 & \textbf{0.0082} & \textbf{0.0068}
 & \textbf{0.0082} & \textbf{0.0067}
 & \textbf{0.0152} & \textbf{0.0073} \\
\bottomrule
\end{tabular}%
}
\end{table}

%% file: sections/c6.ResultsAnalysis/heat_table.tex
\begin{table}[!ht]
\vspace{-2ex}
\small
\caption{Mean Absolute Error (MAE) on 1D convection system for extrapolation tasks ($\beta \in \{40, 50, 60, 70, 80\}$).}
\label{tab:convection_mae_extrapolation}
\centering
\begin{tabular}{lcccccc}
\toprule
\textbf{Method} & 40 & 50 & 60 & 70 & 80 & \textbf{Avg} \\
\midrule
\methodpinnf
    & 0.7433 & 0.7337 & 0.8720 & 0.7671 & 0.8253
    & 0.7883 \\

\methodpinnd~\citep{wang2021understanding}
    & 0.0095 & 0.0148 & 0.0881 & 0.1597 & 0.1699
    & 0.0884 \\

\methodcurriculumn~\citep{krishnapriyan_characterizing_2021}
    & 0.0801 & 0.2593 & 0.3215 & 0.3726 & 0.4072
    & 0.2881 \\

\methodevo~\citep{daw_mitigating_2023}
    & 0.0162 & 0.1477 & 0.3557 & 0.4327 & 0.4151
    & 0.2735 \\

\methodmetapde~\citep{qin2022metapdelearningsolvepdes}
    & 0.1588 & 0.2112 & 0.3757 & 0.3570 & 0.5167
    & 0.3239 \\

\methodhyperlr~\citep{cho2023hypernetwork}
    & 0.4008 & 0.4584 & 0.4912 & 0.5359 & 0.5574
    & 0.4887 \\

\mymethod{} \textbf{(Ours)}
    & \textbf{0.0072} & \textbf{0.0119} & \textbf{0.0137} & \textbf{0.0179} & \textbf{0.0871}
    & \textbf{0.0275} \\
\bottomrule
\end{tabular}
\end{table}

%% file: sections/c6.ResultsAnalysis/helm_table.tex
\begin{table}[!ht]
\vspace{-2ex}
\small
\caption{Mean Absolute Error (MAE) on 2D Helmholtz PDE across varying domain parameters $a_{xy}$.\hide{Baseline methods are shown as rows for clearer comparison.}}
\label{tab:helmholtz_mae_methods_flat}
\centering
\resizebox{\textwidth}{!}{
\begin{tabular}{lccccccc}
\toprule
\textbf{Method} & $a_{44}$ & $a_{45}$ & $a_{55}$ & $a_{46}$ & $a_{56}$ & $a_{66}$ & \textbf{Avg} \\
\midrule
\methodpinnf
    & 0.1135 & 0.2276 & 0.5931
    & 0.3447 & 0.7396 & 1.2283
    & 0.5411 \\

\methodpinnd~\citep{wang2021understanding}
    & 0.1645 & 0.3956 & 0.3185
    & 0.3186 & 0.8444 & 0.4784
    & 0.4200 \\

\methodcurriculumn~\citep{krishnapriyan_characterizing_2021}
    & 0.3080 & 0.1630 & 0.8223
    & 0.1331 & 0.4177 & 0.6520
    & 0.4160 \\

\methodevo~\citep{daw_mitigating_2023}
    & 0.4910 & 0.4192 & 0.5143
    & 0.4786 & 0.4550 & 0.3968
    & 0.4590 \\

\methodmetapde~\citep{qin2022metapdelearningsolvepdes}
    & 0.2473
    & 0.4736
    & 0.5982
    & 0.4954
    & 0.4077
    & 0.4329
    & 0.4425 \\

\methodhyperlr~\citep{cho2023hypernetwork}
    & 0.0613
    & 0.0435
    & 0.1599
    & 0.1570
    & 0.7585
    & 2.9437
    & 0.6874 \\


\mymethod{}\textbf{ (Ours)}
    & \textbf{0.0008} & \textbf{0.0017} & \textbf{0.0022}
    & \textbf{0.0032} & \textbf{0.0074} & \textbf{0.1698}
    & \textbf{0.0307} \\
\bottomrule
\end{tabular}
}
\end{table}

%% file: sections/c6.ResultsAnalysis/ns_table.tex
\begin{table}[!ht]
\small
\vspace{-2ex}
\setlength{\tabcolsep}{5pt}
\caption{Pressure ($p$) prediction errors across Reynolds numbers. Best values per column in bold.}
\label{tab:ns_forward}
\centering
\begin{tabular}{lccc|c}
\toprule
\textbf{Method} & 600 & 800 & 1000 & Avg. \\
\midrule
\methodpinnf
& 0.0183
& 0.1934
& 0.0257
& 0.0791 \\

\methodpinnd~\citep{wang2021understanding}
& 0.0239
& 0.0958
& 0.0353
& 0.0517 \\

\methodcurriculumn~\citep{krishnapriyan_characterizing_2021}
& 0.3011
& 0.2597
& 0.2309
& 0.2639 \\

\methodevo~\citep{daw_mitigating_2023}
& 0.3519
& 0.4966
& 0.2620
& 0.3702 \\

\methodmetapde~\citep{qin2022metapdelearningsolvepdes}
& 0.3172
& 0.2997
& 0.2812
& 0.2994 \\

\methodhyperlr~\citep{cho2023hypernetwork}
& 0.2155
& 0.1253
& 0.1910
& 0.1773 \\

\mymethod{}\textbf{ (Ours)}
& \textbf{0.0164}
& \textbf{0.0065}
& \textbf{0.0058}
& \textbf{0.0096} \\
\bottomrule
\end{tabular}
\end{table}

%% file: sections/c6.ResultsAnalysis/r3.tex
\subsection{(RQ3) What is the effect of invariance encoding and gated layer-wise optimization in \mymethod{}? (Ablation Analysis)} 
\vspace{-1.5ex}
We analyze the effect of \emph{Invariance Encoding} (IE) and \emph{Gated Layer-wise Optimization} (GLO) through ablation variants: 
`\mymethod{} w/o (IE,GLO)' (random initialization) and `\mymethod{} w/o GLO' (MAML-style initialization). 
We focus on the challenging \textbf{2D Navier-Stokes} setting, a multi-physics, highly coupled domain where PINN failures are most pronounced. The ablation analysis is conducted in two stages,  (i) Empirical Evaluation (MAE) (ii) Spectral Evaluation of Absolute Error Fields.
\begin{figure}[t]
   \begin{subfigure}[t]{0.45\textwidth}\
   \centering
  \includegraphics[width=0.6\textwidth]{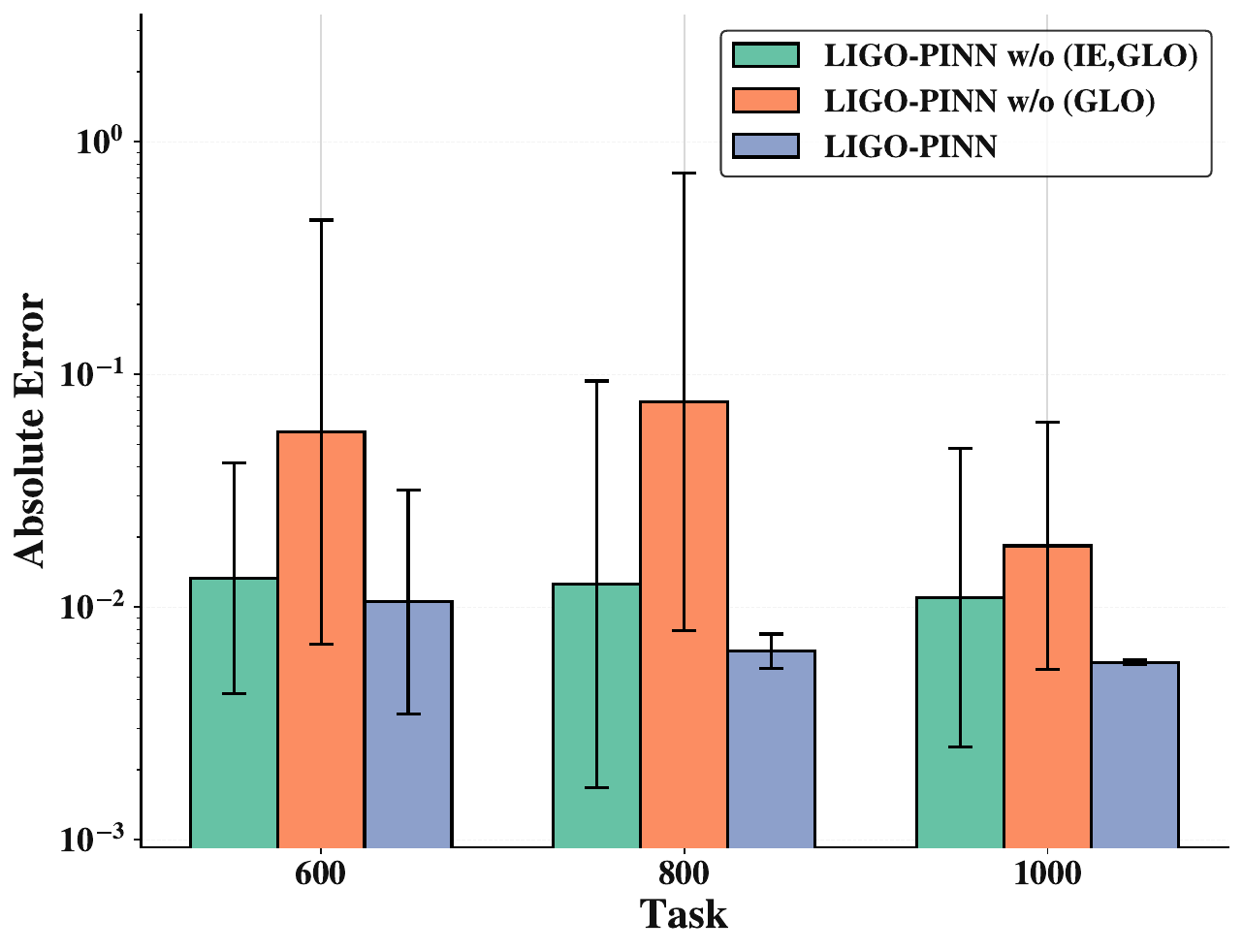}
  \caption{Empirical Evaluation (MAE).}
  \label{fig:ablation_2dns_empirical}
  \end{subfigure}
  \hfill
  \begin{subfigure}[t]{0.45\textwidth}
  \centering
  \includegraphics[width=0.6\textwidth]{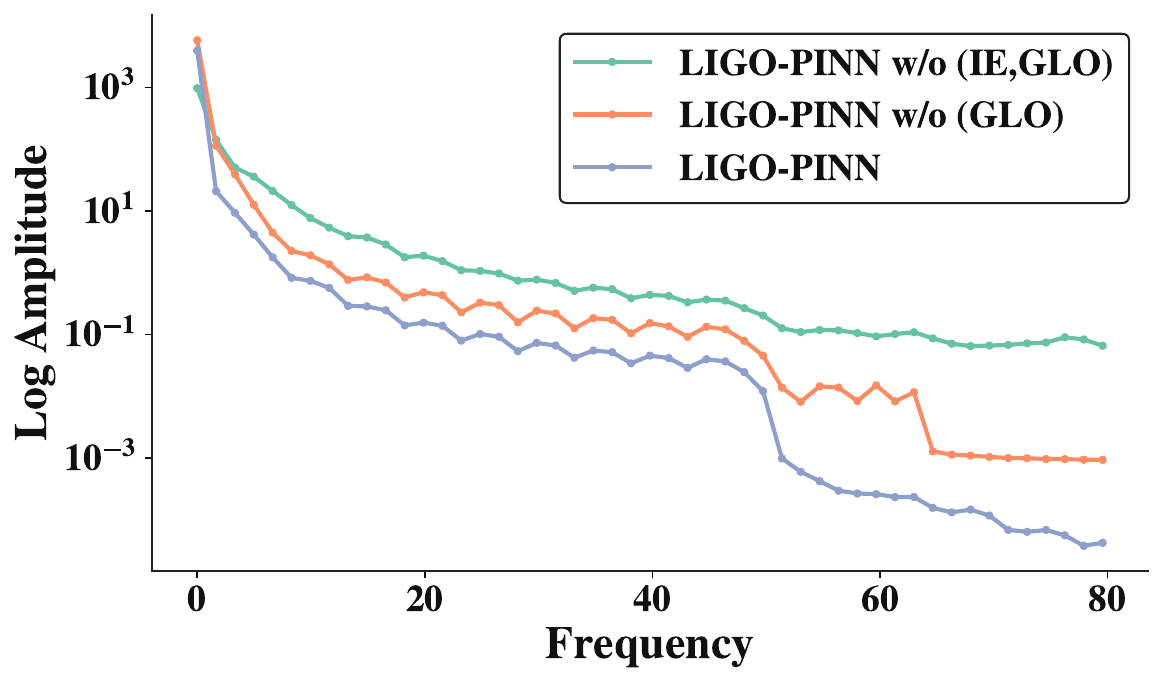}
  \caption{Spectral Evaluation of Error Fields.}
  \label{fig:ablation_2dns_spectral}
  \end{subfigure}
  \vspace{-0.5ex}
  \caption{
  The figure depicts results of \mymethod{} ablation analysis on the most sophisticated PDE domain we evaluated (i.e., 2D Navier-Stokes). Fig.~\ref{fig:ablation_2dns_empirical} depicts empirical (MAE) based ablation characterization for each ablation variant. Fig.~\ref{fig:ablation_2dns_spectral} depicts a spectral distribution of the error fields for each ablation variant (lower is better). Overall, \mymethod{} achieves lower MAE, reduced variance, and mitigates spectral bias relative to ablation variants.}
  \label{fig:ns_ablation}
  \vspace{-4ex}
\end{figure}
\hide{\textbf{(i) Performance Volatility.}  
Fig.~\ref{fig:ns_ablation} (left) shows that random initialization (`w/o (IE,GLO)') exhibits high variance and frequent catastrophic failures. Incorporating IE (`w/o GLO') reduces variance but still results in high error, indicating that invariance encoding alone is insufficient in this multi-physics setting. Full \mymethod{}, with GLO, further lowers both median error and interquartile range, highlighting the critical role of gated, layer-wise parameter optimization in stabilizing training.}
\vspace{-1ex}
\par\noindent
\textbf{Empirical Evaluation.}  
Fig.~\ref{fig:ablation_2dns_empirical} shows that neither simple random initialization (`w/o (IE,GLO)') nor just incorporating IE (`w/o GLO') improves performance on its own. Effective performance is only demonstrated by \mymethod{} where both IE and GLO simultaneously influence learning initialization.
\hide{relative to random initialization but still yields high error, demonstrating that invariance encoding alone is insufficient in this multi-physics setting. Full \mymethod{}, with GLO, achieves substantially lower error,} This highlights the critical role of gated, layer-wise parameter optimization and invariance encoding in learning initial weights that enable stable and accurate convergence in challenging PDE domains.

\vspace{-1ex}
\textbf{Spectral Evaluation.}  
Fig.~\ref{fig:ablation_2dns_spectral}  shows spectral density of the absolute error fields of each ablation variant (lower is better). \mymethod{} consistently achieves lower error across both low and high frequencies, while ablation variants exhibit pronounced high-frequency error (highlighting spectral bias). Specifically, this indicates improved capture of fine-scale structures in fluid dynamics by \mymethod{} compared to variants without IE or GLO. \textit{Results on 1D Convection and 2D Helmholtz exhibit similar trends and detailed in Appendix~\ref{appendix:ablation_additional_pdes}.}
\vspace{-0.5ex}
\par\noindent
\textbf{Additional Results.} We also successfully verify that \mymethod{} performance translates to complex unstructured 3D domains (results in Appendix~\ref{appen:3d_domain_results}). For completeness, we also report computational time and memory footprint details in Appendix~\ref{appendix:computation-resource}.
\hide{further results in Appendix~\ref{appendix:computation-resource}, which details the compute cost of our initialization strategy,} Our results depict that LIGO-PINN scales comparably to standard PINNs aside from a one-time meta-training overhead. Appendix~\ref{appen:sensitivity-analysis} presents sensitivity analyses, to the number and range of sampled tasks during learned initialization phase. 

%% file: sections/c6.ResultsAnalysis/additional.tex

%% file: sections/c7.Conclusion/summary.tex
\vspace{-1ex}
\section{Conclusion}
\vspace{-1ex}
This work shows that \emph{learned initialization} can serve as an effective mechanism for reducing catastrophic failure in PINNs. Rather than relying only on task-specific training heuristics such as curriculum design, adaptive sampling, or loss reweighting, \mymethod{} learns a weight initialization that places the PINN in a more favorable optimization regime before training begins, leading to more stable training and substantially lower errors on challenging extrapolation problems.
Across the evaluated PDE systems, \mymethod{} reduces average error by \textbf{91.5\%} over six comparative PINN baselines and improves over the strongest baseline in each domain by an average of \textbf{81\%}. The ablation results further indicate that both invariance encoding and gated layer-wise optimization are important contributors to this behavior, reducing volatility and improving spectral stability. These findings suggest that learned initialization is a promising direction for more robust PINN training, especially in regimes where standard training often breaks down. Finally, the results on 3D unstructured domains indicate that the benefits of \mymethod{} are not ONLY limited to structured 2D settings and generalize to complex domains.
\vspace{-1ex}
\par\noindent
\textbf{Limitations and Future Work.} While \mymethod{} demonstrates strong performance across several challenging PDE systems,\hide{the current study focuses on a selected set of PDE families and extrapolation regimes commonly used in PINN benchmarking.} the proposed invariance encoding stage requires access to related PDE configurations during initialization. In addition, although the empirical analyses strongly suggest improved optimization behavior and reduced spectral bias, a deeper theoretical understanding of the underlying mechanisms will strengthen the framework and will be explored in the future. Finally, extending \mymethod{} to real-world large-scale multi-physics and industrial simulation settings is a promising avenue for future investigation.

%% file: sections/C9.Appendix/appendix.tex
\appendix
\input{sections/C9.Appendix/A2.PINNformulations}
\input{sections/C9.Appendix/A3.ExperimentalSetup}
\input{sections/C9.Appendix/A4.Additional_results_and_Abalations}

%% file: sections/C9.Appendix/A2.PINNformulations.tex
\begin{center}
    {\LARGE \bf Supplementary LIGO-PINN Experiments\par}
\end{center}
\vspace{1cm}
\section{Detailed formulation of PINN}\label{appendix:pinn_formulation}

In this section, we first present the general formulation of Physics-Informed Neural Networks (PINNs), and then demonstrate its instantiation on two canonical PDEs: the 1D Convection Equation and the 2D Helmholtz Equation.

\subsection{Physics Informed Neural Network}\label{appendix:pinn_primer}

A general form of a PDE can be expressed as:
\begin{equation}
    \mathcal{F}(u(\mathbf{z}); \mathbb{\gamma}) = f(\mathbf{z}), \quad \mathbf{z} \in \Omega,
    \label{eq:physicspde}
\end{equation}
\begin{equation}
    \mathcal{B}(u(\mathbf{z})) = g(\mathbf{z}), \quad \mathbf{z} \in \partial \Omega.
\end{equation}

Here:
\begin{itemize}
    \item $\Omega \subset \mathbb{R}^d$ represents the spatial domain, while $\partial \Omega$ denotes its boundary.
    \item $\mathbf{z} = [x_1, x_2, \dots, x_d, t]^\top \in \mathbb{R}^{d+1}$ encapsulates space-time coordinates.
    \item $u$ is the unknown solution we aim to approximate.
    \item $\mathbb{\gamma}$ denotes physical parameters of the system.
    \item $f(\mathbf{z})$ encodes system-specific data or forcing terms.
    \item $\mathcal{F}$ is a (potentially nonlinear) differential operator characterizing the physical laws of the system.
    \item $\mathcal{B}$ is a boundary operator that enforces initial and boundary conditions, which can be of Dirichlet, Neumann, or periodic types.
\end{itemize}

The PINN framework leverages automatic differentiation to embed the residuals of $\mathcal{F}$ and $\mathcal{B}$ into a composite loss function. This loss function penalizes deviations from the governing equations and boundary conditions, ensuring that the solution adheres to the physical laws while simultaneously fitting observed data.

Neural networks are powerful tools due to their universal approximation property, which allows them to approximate any continuous function given sufficient capacity. In the context of PINNs, we aim to approximate the solution $u(\mathbf{z})$ of the governing equations using a neural network $\hat{u}_\theta(\mathbf{z})$, where $\theta$ denotes the learnable parameters of the neural network. The approximation can be written as:
\begin{equation}
    \hat{u}_\theta(\mathbf{z}) \approx u(\mathbf{z}),
\end{equation}
where $\mathbf{z} \in \mathbb{R}^{d+1}$ represents the space-time coordinates.


To achieve this, the PINN framework constructs a composite loss function consisting of:
\begin{enumerate}
    \item PDE Residual Loss: This enforces the governing physical laws as specified by the PDE.
    \item Boundary/Data Loss: This ensures compliance with boundary or initial conditions, as well as data consistency when available.
\end{enumerate}

\paragraph{PDE Residual Loss}
Using automatic differentiation tools, the neural network computes derivatives efficiently with respect to its inputs, enabling evaluation of the PDE residual. The residual is given by:
\begin{equation}
    \mathcal{R}(\mathbf{z}) = \mathcal{F}(\hat{u}_\theta(\mathbf{z}); \mathbb{\gamma}) - f(\mathbf{z}),
\end{equation}
where $\mathcal{F}$ represents the differential operator, $f(\mathbf{z})$ is the system-specific forcing term, and $\mathbb{\gamma}$ are the physical parameters. The residual loss over the domain $\Omega$ is formulated as:
\begin{equation}
    \mathcal{L}_{{r}} = \frac{1}{N_\Omega} \sum_{i=1}^{N_\Omega} \left| \mathcal{R}(\mathbf{z}_i) \right|^2,
\end{equation}
where $N_\Omega$ represents the number of collocation points in the domain, and $\mathbf{z}_i \in \Omega$ are the sampled points.

\paragraph{Boundary/Data Loss}
For enforcing boundary or initial conditions, we define a loss term that penalizes deviations from the specified boundary values $g(\mathbf{z})$. For Dirichlet conditions, this can be written as:
\begin{equation}
    \mathcal{L}_{{b}} = \frac{1}{N_{\partial \Omega}} \sum_{j=1}^{N_{\partial \Omega}} \left| \mathcal{B}(\hat{u}_\theta(\mathbf{z}_j)) - g(\mathbf{z}_j) \right|^2,
\end{equation}
where $N_{\partial \Omega}$ represents the number of points on the boundary $\partial \Omega$, and $\mathbf{z}_j \in \partial \Omega$ are the boundary points.

\paragraph{Total Loss} The total loss for training the PINN combines the residual loss and the boundary/data loss, weighted by respective coefficients $\lambda_{{r}}$ and $\lambda_{{d}}$:
\begin{equation}
    \mathcal{L}_{{total}} = \lambda_{{r}} \mathcal{L}_{{r}} + \lambda_{{d}} \mathcal{L}_{{b}}.
    \label{eq:pinn_total_loss}
\end{equation}

We calculate the total loss by adding the residual and data losses. Here, $\lambda_{{r}}$ and $\lambda_{{d}}$ are hyperparameters that control the relative contribution of the residual and data losses during training. The residual loss acts as a regularization term, ensuring that the learned solution adheres to physically consistent laws.

To train the neural network, we solve the optimization problem:
\begin{equation}
    \theta^* = \arg\min_{\theta} \mathcal{L}_{{total}}.
    \label{eq:optim_loss}
\end{equation}

We aim to formulate a Physics-Informed Neural Network (PINN) by studying two fundamental classification of systems in physics that are crucial for understanding and modeling the physical world. These systems are categorized based on whether their properties depend on time:

Time-Independent Systems
These systems are used to study the steady-state behavior of a system, where properties such as energy, momentum, or other conserved quantities remain constant over time. Time-independent systems help us analyze static or equilibrium conditions, providing insights into the inherent properties of the system without considering temporal evolution.

Time-Dependent Systems
These systems are used to study how the state of a system evolves over time. By understanding the temporal dynamics, we can gain insights into transient phenomena, system responses, and changes in state variables due to external forces or intrinsic behaviors.

To explore these concepts, we explore the following two systems.

\subsection{1D Convection Equation}

This system represents a classic example of a time-dependent system. 
We consider a one-dimensional convection system to demonstrate the application of Physics-Informed Neural Networks (PINNs). The system models the transport of heat or mass in a medium, where the primary variable $u(x, t)$ evolves over space and time. The spatial domain is $x \in [0, 2\pi]$, and the time domain is $t \in [0, 1]$. This setup allows for studying convection behavior under periodic boundary conditions, often encountered in cyclic systems. This formulation is adapted from  ~\cite{krishnapriyan_characterizing_2021}

The governing partial differential equation (PDE) is given by:
\begin{equation}
    \frac{\partial u}{\partial t} + \beta \frac{\partial u}{\partial x} = 0, \quad x \in \Omega, \, t \in [0, T],
\end{equation}
with the initial condition:
\begin{equation}
    u(x, 0) = h(x), \quad x \in \Omega.
\end{equation}

Here:
\begin{itemize}
    \item $\beta$ is the convection coefficient (physical parameter of system $\gamma$ in \ref{eq:physicspde}),
    \item $h(x)$ specifies the initial condition for $u(x, t)$,
    \item $\Omega$ denotes the spatial domain.
\end{itemize}

To evaluate the performance of PINNs, we generate ground truth data using an analytical solution. For constant $\beta$ and periodic boundary conditions, the analytical solution is:
\begin{equation}
    u_{{analytical}}(x, t) = \mathcal{F}^{-1} \left[ \mathcal{F}(h(x)) e^{-i\beta k t} \right],
\end{equation}
where:
\begin{itemize}
    \item $\mathcal{F}$ and $\mathcal{F}^{-1}$ are the Fourier transform and its inverse, respectively,
    \item $i = \sqrt{-1}$ is the imaginary unit,
    \item $k$ represents the frequency in the Fourier domain.
\end{itemize}

For this problem, we assume:
\begin{equation}
    h(x) = \sin(x), \quad u(0, t) = u(2\pi, t),
\end{equation}
implying periodic boundary conditions.

\paragraph{Residual Loss} The residual loss for this PDE is given by:
\begin{equation}
    \mathcal{L}_{{r}} = \frac{1}{N_f} \sum_{i=1}^{N_f} \left( \frac{\partial \hat{u}}{\partial t} + \beta \frac{\partial \hat{u}}{\partial x} \right)^2,
\end{equation}
where:
\begin{itemize}
    \item $N_f$: Number of collocation points sampled from the interior of the domain $\Omega \times [0, T]$
    \item $\hat{u}$ is the predicted solution,
        \item $\frac{\partial \hat{u}}{\partial t}$: Temporal derivative of the predicted solution,
            \item $\frac{\partial \hat{u}}{\partial x}$: Spatial derivative of the predicted solution,
             \item $\beta$:Convection coefficient which is the parameter describing the physical system

\end{itemize}

\paragraph{Data Loss} The data loss consists of two components: the loss enforcing the initial condition and the loss enforcing the boundary conditions. It is defined as:
\begin{equation}
    \mathcal{L}_{{d}} = \mathcal{L}_{{ic}} + \mathcal{L}_{{bc}},
\end{equation}
where:
\begin{equation}
    \mathcal{L}_{{ic}} = \frac{1}{N_u} \sum_{i=1}^{N_u} \left( \hat{u}(x_i, 0) - g(x_i, 0) \right)^2,
\end{equation}
\begin{equation}
    \mathcal{L}_{{bc}} = \frac{1}{N_b} \sum_{j=1}^{N_b} \left( \hat{u}(0, t_j) - \hat{u}(2\pi, t_j) \right)^2.
\end{equation}

Here:
\begin{itemize}
    \item \( \mathcal{L}_{{ic}} \): Initial condition loss, ensuring the predicted solution \(\hat{u}(x, 0)\) matches the data \(g(x, 0) = h(x)\),
    \item \( \mathcal{L}_{{bc}} \): Boundary condition loss, ensuring periodic boundary conditions \(\hat{u}(0, t) = \hat{u}(2\pi, t)\),
    \item \( N_u \): Number of points sampled from the initial condition,
    \item \( N_b \): Number of points sampled from the boundary condition,
    \item \( g(x_i, 0) = h(x_i) \): Known solution at initial condition points,
    \item \( \hat{u}(0, t_j) \) and \( \hat{u}(2\pi, t_j) \): Predicted solutions at boundary points.
\end{itemize}

\paragraph{Total Loss}
The total loss combines the above components:
\begin{equation}
    \mathcal{L}(\theta) = \lambda_{{r}} \mathcal{L}_{{r}} + \lambda_{{d}} \mathcal{L}_{{d}},
\end{equation}
where:
\begin{itemize}
    \item $\lambda_{{r}}, \lambda_{{d}}$ are weighting coefficients balancing the contributions of the respective terms.
\end{itemize}

\subsection{2D Helmholtz Equation}

We study the vibration states of a two-dimensional membrane using the 2D Helmholtz equation. This equation is widely used to describe stationary wave fields, such as vibrations in membranes, sound waves, and electromagnetic waves. Specifically, we consider a square membrane with spatial coordinates $x_1, x_2 \in [-1, 1]$ and analyze its steady-state behavior under fixed boundary conditions. The Helmholtz equation helps identify resonance patterns and steady-state vibration modes in such systems. The formulation is adapted from~\cite{wang2021understanding}.

The governing PDE for the 2D Helmholtz equation is:
\begin{equation}
    \frac{\partial^2 u}{\partial x_1^2} + \frac{\partial^2 u}{\partial x_2^2} + k^2 u = q(x_1, x_2), \quad (x_1, x_2) \in \Omega,
\end{equation}
where $\Omega = [-1, 1] \times [-1, 1]$ is the spatial domain.

The boundary of the domain, $\partial \Omega$, represents the fixed edges of the membrane. Consequently, the displacement $u(x_1, x_2)$ at the boundaries is zero, giving rise to the Dirichlet boundary condition:
\begin{equation}
    u(x_1, x_2) = 0, \quad (x_1, x_2) \in \partial \Omega.
\end{equation}

We specifically choose $k^2 \neq a_1^2 + a_2^2$ to ensure that the selected $k$ does not correspond to an eigenmode of the domain, resulting in a non-zero source term $q(x_1,x_2)$. Thus, the PDE system is inherently inhomogeneous.

Here:
\begin{itemize}
    \item $k$ is the wavenumber and is kept fixed at $k=1$.
    \item $a_1, a_2 \in \mathbb{Z}^+$ are integer mode numbers representing the spatial oscillation patterns. These serve as the primary PDE parameters affecting solution complexity and difficulty.
    \item $q(x_1, x_2)$ is the resulting source term that characterizes external excitation or forcing within the membrane, driving the system response.
\end{itemize}

For this setup, we use an analytical solution of the form:
\begin{equation}
    u(x_1, x_2) = \sin(a_1 \pi x_1) \sin(a_2 \pi x_2),
\end{equation}
which satisfies the zero-displacement boundary condition. The corresponding source term is derived as:
\begin{equation}
    q(x_1, x_2) = \left[ k^2 - (a_1 \pi)^2 - (a_2 \pi)^2 \right] \sin(a_1 \pi x_1) \sin(a_2 \pi x_2).
\end{equation}

\paragraph{Residual Loss} The residual loss quantifies deviations from the Helmholtz equation at collocation points in $\Omega$:
\begin{equation}
    \mathcal{L}_{\text{residual}} = \frac{1}{N_f} \sum_{i=1}^{N_f} \left( \frac{\partial^2 \hat{u}}{\partial x_1^2} + \frac{\partial^2 \hat{u}}{\partial x_2^2} + k^2 \hat{u} - q(x_{1,i}, x_{2,i}) \right)^2,
\end{equation}
where:
\begin{itemize}
    \item $N_f$: Number of collocation points sampled from the interior of the domain $\Omega$,
    \item $\hat{u}$: Predicted solution obtained from the neural network,
    \item $q(x_{1,i}, x_{2,i})$: Source term evaluated at the collocation point $(x_{1,i}, x_{2,i})$.
\end{itemize}

\paragraph{Boundary Condition Loss} The boundary loss enforces zero displacement at the edges of the membrane, consistent with the fixed boundary condition:
\begin{equation}
    \mathcal{L}_{\text{boundary}} = \frac{1}{N_b} \sum_{i=1}^{N_b} \hat{u}(x_{1,i}, x_{2,i})^2,
\end{equation}
where $N_b$ is the number of points sampled from the boundary $\partial \Omega$.

\paragraph{Total Loss} The total loss combines the residual and boundary condition losses:
\begin{equation}
    \mathcal{L}(\theta) = \lambda_{r} \mathcal{L}_{\text{residual}} + \lambda_{b} \mathcal{L}_{\text{boundary}},
\end{equation}
where $\lambda_{r}$ and $\lambda_{b}$ are weighting coefficients balancing the contributions of the residual and boundary losses.

\subsection{2D Incompressible Navier-Stokes (with learnable coefficients)}

We consider the two-dimensional incompressible Navier-Stokes equations in non-dimensional form, augmented with learnable coefficients for the advection and viscosity terms:
\begin{align}
u_t + \lambda_{\text{adv}} (u u_x + v u_y) &= -\,p_x + \lambda_{\text{vis}} (u_{xx} + u_{yy}), \label{eq:ns_u_learn}\\
v_t + \lambda_{\text{adv}} (u v_x + v v_y) &= -\,p_y + \lambda_{\text{vis}} (v_{xx} + v_{yy}), \label{eq:ns_v_learn}\\
u_x + v_y &= 0. \label{eq:ns_cont_learn}
\end{align}
Here, $u(t,x,y)$ and $v(t,x,y)$ are the velocity components, $p(t,x,y)$ is the pressure, and $\lambda_{\text{adv}}, \lambda_{\text{vis}}$ are trainable scalars. In a \emph{forward} setting, $p$ is predicted by the network but not supervised; in an \emph{inverse} setting, $\lambda_{\text{adv}}$ and $\lambda_{\text{vis}}$ are learned from data. 

\paragraph{Network Outputs}
Let a neural network $\mathcal{N}_\theta(t,x,y)$ predict
\[
(\hat{u}, \hat{v}, \hat{p}) = \mathcal{N}_\theta(t,x,y),
\]

\paragraph{Residual Loss}
Define the PDE residuals at collocation points $\{(t_i,x_i,y_i)\}_{i=1}^{N_f}$ by
\begin{align}
f_u &= \hat{u}_t + \lambda_{\text{adv}} (\hat{u}\,\hat{u}_x + \hat{v}\,\hat{u}_y) + \hat{p}_x - \lambda_{\text{vis}} (\hat{u}_{xx} + \hat{u}_{yy}), \\
f_v &= \hat{v}_t + \lambda_{\text{adv}} (\hat{u}\,\hat{v}_x + \hat{v}\,\hat{v}_y) + \hat{p}_y - \lambda_{\text{vis}} (\hat{v}_{xx} + \hat{v}_{yy}), \\
f_c &= \hat{u}_x + \hat{v}_y ,
\end{align}
and the residual loss
\begin{equation}
    \mathcal{L}_{r} \;=\; \frac{1}{N_f} \sum_{i=1}^{N_f} 
    \Big( \,|f_u(t_i,x_i,y_i)|^2 \;+\; |f_v(t_i,x_i,y_i)|^2 \;+\; \alpha\,|f_c(t_i,x_i,y_i)|^2 \Big),
\end{equation}
where $\alpha>0$ weights the continuity residual.

\paragraph{Data Loss (Vortex Shedding Velocities)}
Given velocity measurements from the vortex-shedding flow,
\(
\{(t_i,x_i,y_i,u_i,v_i)\}_{i=1}^{N_d},
\)
we use only velocity supervision:
\begin{equation}
    \mathcal{L}_{d} \;=\; \frac{1}{N_d} \sum_{i=1}^{N_d}
    \Big( \,(\hat{u}(t_i,x_i,y_i) - u_i)^2 \;+\; (\hat{v}(t_i,x_i,y_i) - v_i)^2 \,\Big).
\end{equation}
Pressure is \emph{not} supervised and is recovered implicitly via the momentum equations.

\paragraph{Total Loss}
The training objective combines residual and data terms:
\begin{equation}
    \mathcal{L}(\theta,\lambda_{\text{adv}},\lambda_{\text{vis}}) 
    \;=\; \lambda_{r}\,\mathcal{L}_{r} \;+\; \lambda_{d}\,\mathcal{L}_{d},
\end{equation}
with weights $\lambda_r,\lambda_d>0$. For inverse problems, $\lambda_{\text{adv}}$ and $\lambda_{\text{vis}}$ are optimized jointly with $\theta$; for forward problems they are fixed (e.g., $\lambda_{\text{adv}}=1$, $\lambda_{\text{vis}}=1/\mathrm{Re}$).

%% file: sections/C9.Appendix/A3.ExperimentalSetup.tex
\section{Experimental Setup}\label{appendix:experimental-setup}

\subsection{Dataset and Task Construction}
\label{sec:dataset_desc}

We evaluate our method across three PDE systems: 1D Convection, 2D Helmholtz, and 2D Navier-Stokes. 
For each PDE system, tasks are constructed by varying the governing PDE parameters following the experimental protocols established in prior PINN benchmark works.

\paragraph{1D Convection.}
For the 1D Convection system, we use the analytical solution formulation introduced in \methodcurriculumn{}~\cite{krishnapriyan_characterizing_2021}. 
The dataset generation and task construction closely follow the publicly available benchmark implementation provided in their codebase. 
Task difficulty is controlled through the convection coefficient $\beta$, where each task corresponds to a different value of $\beta$.

\paragraph{2D Helmholtz.}
For the 2D Helmholtz system, we construct our own dataset using the analytical solution formulation described in~\cite{wang2021understanding}. 
The source term is analytically derived from the selected closed-form solution, allowing exact supervision of PDE behavior. 
Task construction is performed by varying the spatial frequency parameters $(a_1,a_2)$, which control the oscillatory complexity of the PDE solution.

\paragraph{2D Navier-Stokes.}
For the 2D Navier-Stokes system, we generate fluid simulation data using \texttt{PhiFlow}, a differentiable fluid simulation framework based on finite element style numerical solvers. 
The simulation setup follows the cylinder wake configuration commonly adopted in PINN literature and is designed to closely match the benchmark setting introduced by~\cite{raissi2017physicsinformeddeeplearning},. 
Task difficulty is controlled by varying the Reynolds number ($Re$), where higher Reynolds regimes correspond to increasingly complex flow dynamics.

Across all PDE systems, task families are constructed by systematically varying the corresponding PDE parameters, following the difficulty progression protocols established in the aforementioned baseline works.

\subsection{Task Difficulty Characterization via PINN Loss Dynamics}
\label{app:task_difficulty_boxplots}
In many PDE settings, easy vs. hard regimes are \emph{known} and hence task difficulty is defined with the aid of domain expertise, where practitioners identify regimes that are known to be easy or challenging based on the underlying physics. However, in the absence of such prior knowledge, purely empirical criteria can be adopted, based on PINN training dynamics. Specifically, we observe that the distribution of PINN training losses even after early-stage training (i.e., 10K epochs) already provides a clear and consistent signal of task difficulty. Across PDE systems, this behavior exhibits a distinct transition in error dynamics, allowing us to separate regimes where standard PINN training is effective from those where it begins to fail.
\begin{figure}[h]
    \centering
    \begin{subfigure}[t]{0.48\textwidth}
        \centering
        \includegraphics[width=\textwidth]{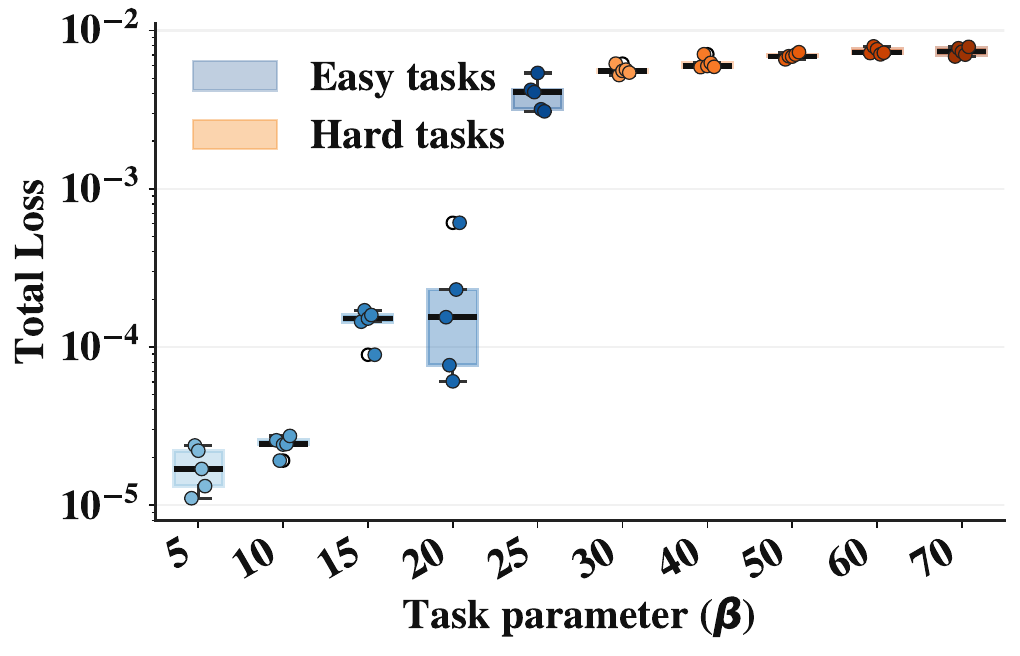}
        \caption{1D Convection}
        \label{fig:heat_boxplot}
    \end{subfigure}
    \hfill
    \begin{subfigure}[t]{0.48\textwidth}
        \centering
        \includegraphics[width=\textwidth]{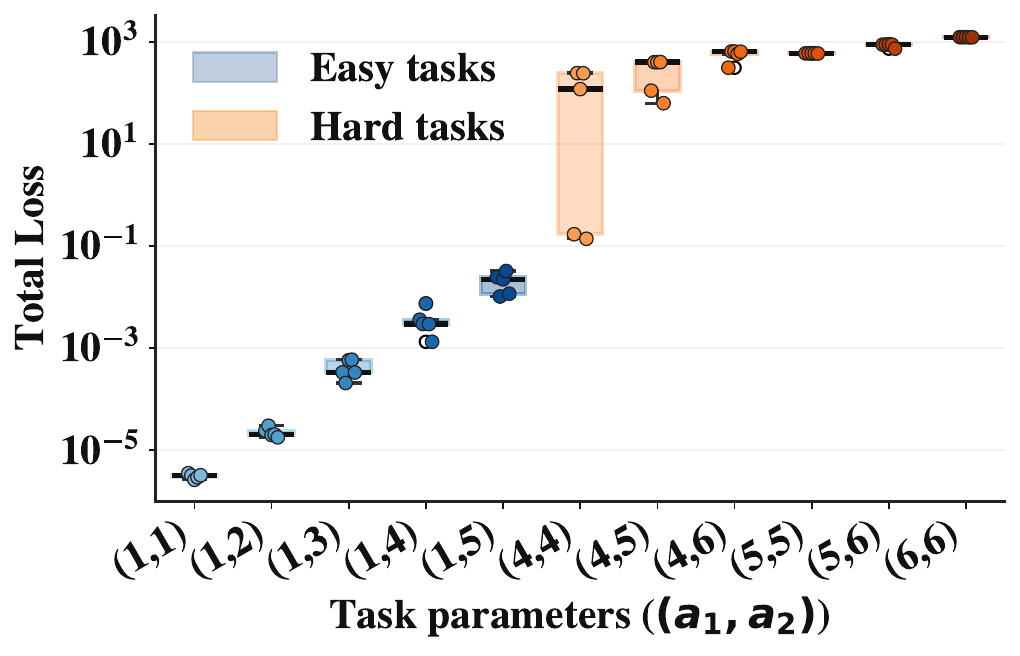}
        \caption{2D Helmholtz}
        \label{fig:helm_boxplot}
    \end{subfigure}
    \caption{
    Distribution of residual losses @ 10K across five random seeds. 1D Convection exhibits a gradual transition in difficulty as $\beta$ increases, whereas Helmholtz shows a significantly sharper transition due to increasing oscillatory complexity.}
    \label{fig:residual_boxplots_combined}
\end{figure}
\paragraph{1D Convection.}
For the convection equation, task complexity is governed by the coefficient $\beta$. As shown in Fig.~\ref{fig:residual_boxplots_combined}(a), lower values of $\beta$ yield consistently low training loss across random seeds, indicating stable optimization. As $\beta$ increases, the magnitude of the  loss grows, with a sharp increase in magnitude at the transition from easy to difficult task boundary.\hide{We observe a gradual transition in behavior, with $\beta \approx 25$ lying near the boundary between easy and hard regimes.} Note, despite $\beta=25$ exhibiting `hard task' dynamics, it is included in the easy task set to maintain consistency with the experimental setup employed in  \methodcurriculumn (~\cite{krishnapriyan_characterizing_2021}) a state-of-the-art baseline we compare with in our work.
Based on this transition, we consider $\beta \geq 30$ as the onset of the hard regime, and treat all tasks with $\beta \geq 30$ as hard tasks in our experiments.

\paragraph{2D Helmholtz.}
In the 2D Helmholtz equation, task complexity is controlled by the parameters $(a_1, a_2)$ which govern the domain spatial frequency (given by $\sqrt{a_1^2 + a_2^2}$). As shown in Fig.~\ref{fig:residual_boxplots_combined}(b), low-frequency configurations produce low and stable PINN training losses, indicating that PINNs can effectively learn task dynamics. In contrast, hard-task settings (i.e., high frequency configurations) exhibit a sharp increase in loss magnitude\hide{as well as greater residual loss variability across seeds}.\hide{Notably, even moderate-frequency configurations such as $(4,4)$ already show significant degradation in optimization performance, indicating a much earlier and sharper transition to the hard regime compared to the convection system.}

\subsection{Baseline Model Description}\label{sec:model_explain}
We now provide brief descriptions of baseline models employed for comparative evaluation.

\begin{enumerate}
\item \methodpinnf{} Raissi Original Formulation
\item \methodpinnd{}~\cite{wang2021understanding,wu2023comprehensive}: The current optimized and stable PINN training paradigm. In each iteration, collocation points are randomly resampled using quasi-random low-discrepancy sequences, combined with adaptive learning rate scheduling to improve convergence.
    \item \methodcurriculumn{}~\cite{krishnapriyan_characterizing_2021}: Curriculum regularization of PINN which progressively trains models from easier to harder PDE tasks by incrementally introducing complexity during training. If we want to Learn a 1D convection system with $\beta=30$. We incremently train on $\beta=\{5,10,15,20,25,30\}$
    
    \item \methodevo{} ~\cite{daw_mitigating_2023}:
    Employs the adaptive sampling strategy Retain-Resample-Release (R3), focusing training effort by retaining collocation points with high residual errors while periodically resampling the remaining points.
    \item \methodmetapde{} ~\cite{qin2022metapdelearningsolvepdes}:
    A meta-learning framework for PDEs based on a MAML-style formulation. It also includes a learnable per-parameter step-size.
    \item \methodhyperlr~\cite{cho2023hypernetwork}:
    A hypernetwork-based meta-learning method that generates low-rank, task-specific PINN weights for parameterized PDEs.
    \item \mymethod{} w/o (IE, GLO): Effectively standard PINN training from randomly initialized weights.
    \item \mymethod{} w/o GLO: An adaptation of the Model-Agnostic Meta-Learning–based framework.
\end{enumerate}

\subsection{Training Configuration per PDE System}
\label{appen:exp-setup-conv1d}

We summarize the paper-facing experimental configurations used across all PDE systems.
Unless otherwise specified, all models use Xavier weight initialization.
Standard PINN-style baselines are optimized with Adam and cosine annealing learning-rate scheduling.
For fair comparison, baseline methods and \mymethod{} share the same network architectures, task definitions, collocation budgets, and evaluation protocols within each PDE system.
The training horizon is matched to the method family.
One-phase baselines are trained for 56{,}000 steps.
LIGO/MAML-based downstream fine-tuning is run for 50{,}000 steps.

For \mymethod{}, the LIGO invariance encoding stage is meta-trained exclusively on the easy-task distribution $\Gamma_{\mathrm{Easy}}$, while evaluation is conducted on the unseen hard-task distribution $\Gamma_{\mathrm{Hard}}$.
During invariance encoding, the model learns shared initialization priors across related PDE tasks.
It is then adapted to harder PDE regimes.
Subsequently, gated layer-wise optimization (GLO) progressively adapts deeper network layers during task-specific fine-tuning.
The corresponding outer and inner optimization steps and learning rates are reported explicitly in the tables below.

Unless explicitly stated otherwise, all reported MAE results are averaged across five random seeds.
Within each PDE system, the same collocation point budgets, boundary-condition definitions, and evaluation protocols are used across competing methods.
Table~\ref{tab:conv-exp-setup} reports the settings for the 1D Convection system,
Table~\ref{tab:helmholtz-exp-setup} reports the settings for the 2D Helmholtz system,
and Table~\ref{tab:ns-exp-setup} reports the settings for the 2D Navier-Stokes system.

\begin{table}[htbp]
\centering
\small
\caption{1D Convection Experimental Settings}
\label{tab:conv-exp-setup}
\begin{tabularx}{\linewidth}{lX}
\toprule
\multicolumn{2}{l}{\textbf{Common Settings}} \\
\midrule
Equation & $\partial_t u + \beta\,\partial_x u = 0$ \\
Input/Domain & $(x,t)\in[0,2\pi]\times[0,1]$ \\
Dataset Source & Analytical solution adapted from~\cite{krishnapriyan_characterizing_2021} \\
Boundary Condition & Periodic \\
Baseline Sampling & Static and dynamic baselines use matched collocation budgets. \\
Adaptive Sampling & R3 uses adaptive residual-based resampling. \\
Network Architecture & 5 hidden layers, 50 neurons each, \texttt{Tanh}, Xavier init. \\
MLP Parameters & 10{,}401 \\
Collocation Points & $N_f=1000$ \\
Boundary Points & $N_b=1000$ \\
Supervised Data Usage & Matched across compared methods. \\
Training Tasks ($\Gamma_{\mathrm{Easy}}$) & $\beta\in\{5,10,15,20,25\}$ \\
Evaluation Tasks ($\Gamma_{\mathrm{Hard}}$) & $\beta\in\{,40,50,60,70,80\}$ \\

\midrule
\multicolumn{2}{l}{\textbf{Fine-Tuning Settings}} \\
\midrule
Optimizer & Adam \\
Base Learning Rate & 0.01 \\
Scheduler & Cosine scheduling \\
Residual/Data Weight Ratio & $\lambda_{\mathrm{res}}:\lambda_{\mathrm{data}} = 0.01:1$ \\
Layer Gating During Evaluation & False \\
Baseline Training Steps & 56{,}000 \\
LIGO/MAML Fine-Tuning Steps & 50{,}000 \\

\midrule
\multicolumn{2}{l}{\textbf{LIGO Settings}} \\
\midrule
LIGO Outer Steps & 200 \\
LIGO Inner Steps & 10 \\
Layer Gating & True \\
LIGO Outer Optimizer & Adam \\
LIGO Inner Optimizer & SGD \\
LIGO Outer LR & 0.01 \\
LIGO Inner LR & 0.005 \\
\bottomrule
\end{tabularx}
\end{table}

\textbf{Reproducibility Notes.} We provide the supplementary material needed to reproduce the reported results.
Above, we specify the architecture, optimizer, learning-rate, task split, sampling, and training-horizon details used for each PDE system.
The supplementary code package includes README-based instructions for environment setup, experiment execution, and result reproduction.
We also provide the selected model runs and random-seed settings used for the reported evaluations so that the main benchmark tables and figures can be reproduced consistently.

Unless otherwise noted, all final reported MAE values are averaged over five random seeds.
Analytical-task PDE systems are reproduced through the provided task-generation code, while the Navier--Stokes experiments use the supplied simulation/data-generation pipeline derived from the \texttt{PhiFlow} framework.
Experiment metadata, metrics, and artifacts are tracked with MLflow for auditability and result verification.

\begin{table}[htbp]
\centering
\small
\caption{2D Helmholtz Experimental Settings}
\label{tab:helmholtz-exp-setup}
\begin{tabularx}{\linewidth}{lX}
\toprule
\multicolumn{2}{l}{\textbf{Common Settings}} \\
\midrule
Equation & $\Delta u + k^2 u = q(x_1,x_2)$,\; $k=1$ \\
Input/Domain & $(x_1,x_2)\in[-1,1]^2$ \\
Dataset Source & Analytical solution adapted from~\cite{wang2021understanding} \\
Boundary Condition & Dirichlet \\
Baseline Sampling & Static and dynamic baselines use Halton-based sampling. \\
Adaptive Sampling & Adaptive baselines use residual-guided updates under the same collocation budget. \\
Spatial Frequency Parameters & $(a_1,a_2)$ \\
Network Architecture & 5 hidden layers, 50 neurons each, \texttt{Tanh}, Xavier init. \\
MLP Parameters & 10{,}401 \\
Collocation Points & $N_f=1000$ \\
Boundary Points & Matched across compared methods \\
Supervised Data Usage & Matched across compared methods. \\
Training Tasks ($\Gamma_{\mathrm{Easy}}$) & $(a_1,a_2)\in\{(1,1),(1,2),(1,3),(1,4),(1,5)\}$ \\
Evaluation Tasks ($\Gamma_{\mathrm{Hard}}$) & $(a_1,a_2)\in\{(4,4),(4,5),(5,5),(4,6),(5,6),(6,6)\}$ \\

\midrule
\multicolumn{2}{l}{\textbf{Fine-Tuning Settings}} \\
\midrule
Optimizer & Adam \\
Base Learning Rate & 0.01 \\
Scheduler & Cosine scheduling \\
Residual/Data Weight Ratio & $\lambda_{\mathrm{res}}:\lambda_{\mathrm{data}} = 0.01:1$ \\
Layer Gating During Evaluation & False \\
Baseline Training Steps & 56{,}000 \\
LIGO/MAML Fine-Tuning Steps & 50{,}000 \\

\midrule
\multicolumn{2}{l}{\textbf{LIGO Settings}} \\
\midrule
LIGO Outer Steps & 200 \\
LIGO Inner Steps & 10 \\
Layer Gating & True \\
LIGO Outer Optimizer & Adam \\
LIGO Inner Optimizer & SGD \\
LIGO Outer LR & 0.005 \\
LIGO Inner LR & 0.001 \\
\bottomrule
\end{tabularx}
\end{table}

\begin{table}[htbp]
\centering
\small
\caption{2D Navier-Stokes Experimental Settings (cylinder wake)}
\label{tab:ns-exp-setup}
\begin{tabularx}{\linewidth}{lX}
\toprule
\multicolumn{2}{l}{\textbf{Common Settings}} \\
\midrule
Equation & Incompressible NS with learnable $\lambda_{\text{adv}},\ \lambda_{\text{vis}}$ \\
Input/Domain & $(x,y,t)\in \Omega\times[0,T]$ \\
Spatial Geometry & $\Omega$ is a channel with a unit-diameter cylinder \\
Dataset Source & \texttt{PhiFlow} fluid simulation framework \\
Reference Benchmark & Cylinder wake setup adapted from~\cite{raissi2017physicspart1} \\
Baseline Sampling & Fixed/static and dynamic settings use matched collocation budgets. \\
Adaptive Sampling & Adaptive settings use matched budgets with residual-guided resampling. \\
Network Architecture & 7 hidden layers, 50 neurons each, \texttt{Tanh}, Xavier init. \\
MLP Parameters & 15{,}602 \\
Collocation Points & $N_f=5000$ \\
Training Tasks ($\Gamma_{\mathrm{Easy}}$) & $\mathrm{Re}\in\{100,200,300,400,500\}$ \\
Evaluation Tasks ($\Gamma_{\mathrm{Hard}}$) & $\mathrm{Re}\in\{600,800,1000\}$ \\

\midrule
\multicolumn{2}{l}{\textbf{Fine-Tuning Settings}} \\
\midrule
Optimizer & Adam \\
Base Learning Rate & 0.01 \\
Scheduler & Cosine scheduling \\
Residual/Data Weight Ratio & $\lambda_{\mathrm{res}}:\lambda_{\mathrm{data}} = 1:1$ \\
Layer Gating During Evaluation & False \\
Baseline Training Steps & 56{,}000 \\
LIGO/MAML Fine-Tuning Steps & 50{,}000 \\

\midrule
\multicolumn{2}{l}{\textbf{LIGO Settings}} \\
\midrule
LIGO Outer Steps & 200 \\
LIGO Inner Steps & 10 \\
Layer Gating & True \\
LIGO Outer Optimizer & Adam \\
LIGO Inner Optimizer & SGD \\
LIGO Outer LR & 0.001 \\
LIGO Inner LR & 0.005 \\
\bottomrule
\end{tabularx}
\end{table}

\subsection{Computational Resources}
\label{appendix:computation-resource}

All experiments were conducted on an internal Linux GPU server using a four NVIDIA RTX A6000 GPU with 48\,GB vRAM per GPU.
To improve reproducibility, we report the hardware and software environment, the experiment execution setup, and representative runtime and memory costs for the main benchmark pipeline.
The reported specifications correspond to the environment used for the paper-facing experiments.

\small
\begin{table}[ht]
  \centering
  \caption{Hardware and software specifications used for the reported experiments.}
  \label{tab:comp_resources_specs}
  \begin{tabular}{|l|l|}
    \hline
    \textbf{Component} & \textbf{Specification} \\ \hline
    Compute Platform & Internal Linux GPU server \\ \hline
    GPU & NVIDIA RTX A6000 (48\,GB vRAM) \\ \hline
    CPU & Intel(R) Xeon(R) Platinum 8358 CPU @ 2.60\,GHz \\ \hline
    System RAM & 2\,TB DDR4 \\ \hline
    Operating System & Ubuntu 20.04 LTS \\ \hline
    Python & 3.10 \\ \hline
    PyTorch & 2.4 \\ \hline
    CUDA & 12 \\ \hline
    Experiment Tracking & MLflow 2.14 \\ \hline
    Memory Monitoring & \texttt{nvidia-smi} \\ \hline
  \end{tabular}
\end{table}
\normalsize

\paragraph{Computation Cost.}
To contextualize the computational overhead of invariance encoding, we report representative wall-clock time (minutes) and peak GPU memory usage (MB) in Table~\ref{tab:computation_cost_total}.
All measurements were taken on the same NVIDIA RTX A6000 GPU to ensure consistency across stages.
Wall-clock time was recorded per run, and peak GPU memory was measured using \texttt{nvidia-smi}.
The table reports representative benchmark costs for the 1D Convection system and should be interpreted as stage-level execution cost rather than full cumulative project compute.

The overhead of \mymethod{} is defined as the additional runtime introduced by the learned initialization meta-learning stage relative to PINN-D training. Across the three PDE systems, PINN-D requires 51.82 GPU-hours, while \mymethod{} requires 62.57 GPU-hours when both learned initialization via meta-learning and downstream fine-tuning are included. This corresponds to an additional 10.75 GPU-hours, or a 20.75\% cumulative runtime overhead. This overhead is incurred during the initialization stage and can be amortized when the learned initialization is reused across multiple downstream tasks within the same PDE family.

The higher peak memory observed for the Navier--Stokes system is primarily due to the larger dataset and the additional automatic-differentiation graph required by the coupled, higher-order PDE residual terms. Peak memory can be reduced by using a first-order MAML-style update, where higher-order gradients through the inner-loop adaptation are not tracked. This would improve runtime and memory efficiency, but it trades off meta-gradient fidelity and may reduce final accuracy. In this work, we keep the higher-order meta-gradient computation and use the same inner-loop setting across PDE systems to maintain experimental parity.
\begin{table}[ht]
\scriptsize
\setlength{\tabcolsep}{4pt}
\caption{Cumulative computation cost across PDE systems. PINN-D reports target-domain training cost across all evaluation tasks and five random seeds. \mymethod{} reports the total cost, including cost of learned initialization via meta-learning and downstream fine-tuning. Overhead is computed relative to PINN-D. Peak memory is reported as the maximum observed GPU memory and is not summed across runs.}
\label{tab:computation_cost_total}
\centering
\begin{tabular}{@{}lrrrrrrr@{}}
\toprule
\textbf{PDE}
& \textbf{Runs}
& \multicolumn{2}{c}{\textbf{PINN-D}}
& \multicolumn{2}{c}{\textbf{\mymethod{} Total}}
& \multicolumn{2}{c}{\textbf{Overhead}} \\
\cmidrule(lr){3-4}
\cmidrule(lr){5-6}
\cmidrule(l){7-8}
&
& \textbf{Time (h)} & \textbf{GB}
& \textbf{Time (h)} & \textbf{GB}
& \textbf{Time (h)} & \textbf{\%} \\
\midrule
1D Convection
& 25
& 9.73 & 2.10
& 10.82 & 2.30
& 1.09 & 11.23 \\

2D Helmholtz
& 30
& 21.56 & 0.70
& 22.88 & 1.12
& 1.32 & 6.10 \\

2D Navier--Stokes
& 15
& 20.53 & 1.70
& 28.87 & 19.12
& 8.34 & 40.64 \\
\midrule
\textbf{Total}
& \textbf{70}
& \textbf{51.82} & \textbf{2.10}
& \textbf{62.57} & \textbf{19.12}
& \textbf{10.75} & \textbf{20.75} \\
\bottomrule
\end{tabular}
\end{table}
The invariance-encoding stage is a one-time cost.
Once learned, the initialization can be reused across multiple tasks within the same PDE family, thereby amortizing the additional meta-training overhead.
During the reported comparisons, the collocation budgets and evaluation settings were kept fixed between invariance encoding, downstream fine-tuning, and standard PINN training.
This shows that \mymethod{} retains comparable per-task fine-tuning cost to standard PINNs, with the main additional expense arising from the one-time initialization stage.

%% file: sections/C9.Appendix/A4.Additional_results_and_Abalations.tex
\section{Additional Results}\label{appendix:extended_results}

In this section, we present supplementary training results to further evaluate the behavior of our methods.

\vspace{-0.5ex}
\input{sections/c6.ResultsAnalysis/r4}

\subsection{Hessian-Based Loss-Landscape Visualization}\label{appendix_hessian_helm}
To better understand the local geometry of the trained PINN, we analyze the curvature of each loss component—total loss, data loss, and residual loss—around the converged weights~$\theta^*$.  Concretely:

\begin{figure}[ht]
  \centering
  \begin{subfigure}[t]{0.9\textwidth}
    \centering
    \includegraphics[width=\textwidth]{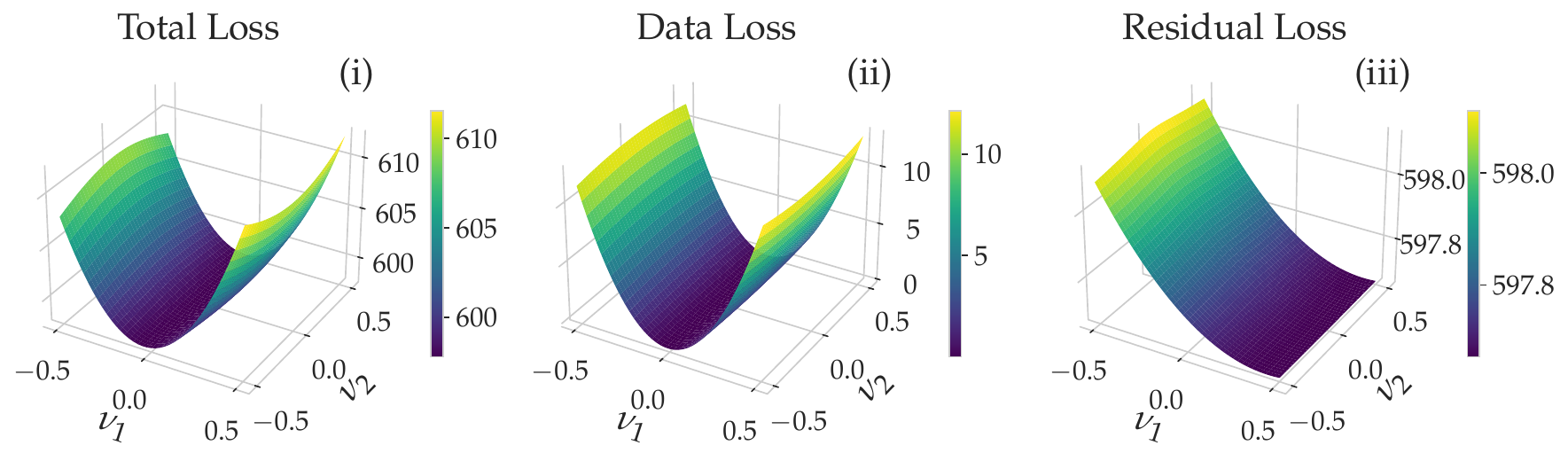}
    \caption{2D Helmholtz PINN Loss Landscape}
    \label{fig:helm2d_xavier_surface}
  \end{subfigure}
  \vfill
  \begin{subfigure}[t]{0.9\textwidth}
    \centering
    \includegraphics[width=\textwidth]{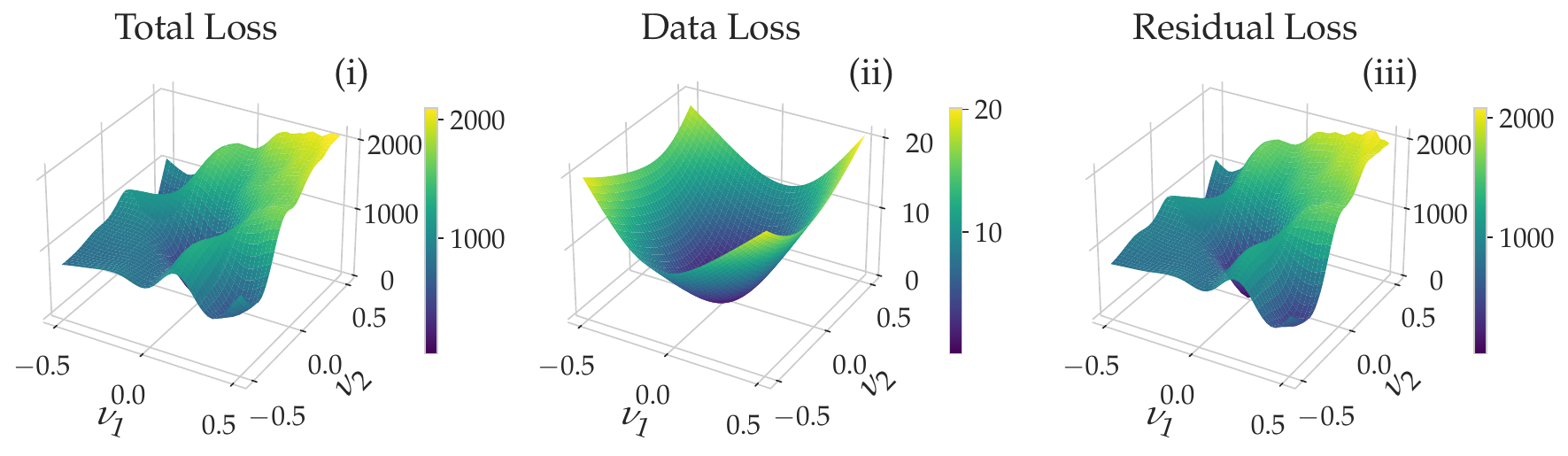}
    \caption{2D Helmholtz \mymethod{} Loss Landscape}
    \label{fig:helm2d_LIGOpinn_surface}
  \end{subfigure}

  \caption[PINN Spectral Surfaces]{%
    Loss Landscape visualization by pertubating model along its top-2 eigenvectors ($\nu_{1}$,$\nu_{2}$) plots for PINN Training (Total, Data, Residuals) in the two case of PINN in 2D Helmholtz System($a_{55}$):
    (a) Randomly Initialized PINN, (b) \mymethod{}.
  }
  \label{fig:appendix_helm_hessian_plot}
\end{figure}

\begin{enumerate}
  \item \textbf{Calculate losses.}  Let
  \[
    L_{\mathrm{tot}}(\theta),\quad 
    L_{\mathrm{data}}(\theta),\quad 
    L_{\mathrm{res}}(\theta)
  \]
  denote the total, data, and residual losses, respectively, as in Eq.~(2).
  \item \textbf{Compute Hessians.}  For each loss $L\in\{L_{\mathrm{tot}},L_{\mathrm{data}},L_{\mathrm{res}}\}$, form the Hessian
  \[
    H_L = \nabla^2_\theta\,L(\theta)\bigl\lvert_{\theta=\theta^*}\,.
  \]
  \item \textbf{Eigen-decomposition.}  Solve
  \[
    H_L\,v_i = \lambda_i\,v_i,\quad
    \lambda_1 \ge \lambda_2 \ge \dots,
  \]
  and retain the top two eigenpairs $(\lambda_1,v_1)$, $(\lambda_2,v_2)$, which capture the directions of greatest curvature.
  \item \textbf{Parameter perturbation.}  For offsets $(\alpha,\beta)\in[-\delta,\delta]^2$, define
  \[
    \theta(\alpha,\beta) \;=\; \theta^* \;+\;\alpha\,v_1 \;+\;\beta\,v_2\,.
  \]
  \item \textbf{Landscape slice.}  The two-dimensional landscape
  \[
    f_L(\alpha,\beta) \;=\; L\bigl(\theta(\alpha,\beta)\bigr)
  \]
  is evaluated on a uniform grid of $(\alpha,\beta)$ and visualized as a heatmap.
\end{enumerate}

\textbf{2D Helmholtz Loss-Landscape Visualization.}
In \ref{fig:helm2d_xavier_surface} we see that in case that model has converge to a trivial solution that has trapped in steep 1 dimension wells and stops which does nto faciliates learning.
on the other hand \ref{fig:helm2d_LIGOpinn_surface} we that a converged model has smoother optimization bowls that facilates good convergence

\input{sections/C9.Appendix/AE5.effect_of_initialization}

\subsection{Compatibility with Existing PINN Stabilization Methods}
\label{appen:additional-results}

We further investigate whether the proposed learned initialization strategy is compatible with existing PINN stabilization approaches. To this end, we integrate our learned initialization into the \methodevo{} sampling framework (R3)~\cite{daw_mitigating_2023}, a state-of-the-art adaptive collocation sampling method designed to improve PINN convergence stability.

Specifically, we compare two variants of the \methodevo{} framework:

\begin{itemize}
    \item \textbf{\methodevo{} (Xavier Init.)}: standard \methodevo{} training initialized using Xavier initialization,
    \item \textbf{\methodevo{} (LIGO Init.)}: \methodevo{} initialized using learned weights obtained from our invariance encoding setup.
\end{itemize}

To provide a stronger stress test, both variants were trained for \textbf{100K epochs}, significantly extending beyond the primary training budget used throughout this work. Evaluation was performed on challenging extrapolation regimes of the 1D Convection system with $\beta \in \{40,50,60\}$. For fairness, both models used identical architectures, optimization schedules, and collocation settings. Results are averaged across \textbf{5 independent random seeds}, and we report both mean MAE and standard deviation.

As shown in Table~\ref{tab:adapt-mae}, the proposed learned initialization remains compatible with the \methodevo{} framework even under substantially extended training schedules. In particular, the learned initialization achieves lower MAE than Xavier initialization in the $\beta=40$ and $\beta=60$ extrapolation settings, including a substantial improvement in the more challenging $\beta=60$ regime. These results suggest that the proposed initialization mechanism is complementary to existing PINN stabilization strategies and can be integrated into adaptive sampling frameworks in a model-agnostic manner.

\begin{table}[!ht]
\small
\centering
\caption{Comparison of \methodevo{} performance on the 1D Convection system under extended 100K epoch training. Results are reported as mean MAE across training runs}
\label{tab:adapt-mae}
\begin{tabular}{lcc}
\toprule
\textbf{1D Convection ($\beta$)} & \textbf{\methodevo{} (Xavier Init.)} & \textbf{\methodevo{} (LIGO Init.)} \\
\midrule
40 & 0.0084 & 0.0066 \\
50 & 0.0146 & 0.0105 \\
60 & 0.2086 & 0.0228 \\
\midrule
\textbf{Average MAE} & \textbf{0.0772} & \textbf{0.0133} \\
\bottomrule
\end{tabular}
\end{table}


\subsection{Evaluation across baseline}\label{appen:baseline_tables}

\paragraph{1D Convection.}
For the 1D convection system, we evaluate all methods under progressively harder extrapolation regimes with convection coefficients $\beta \in \{40,50,60,70,80\}$. 
In addition to the mean prediction error, Table~\ref{tab:1dconv_mean_stdev_extrapolation} also reports the variance across independent runs to characterize optimization stability under increasing transport difficulty. 
As the convection coefficient increases, standard PINN variants exhibit larger performance degradation and higher variance, indicating unstable convergence dynamics. 
In contrast,~\mymethod{} maintains consistently low error and reduced variance across most extrapolation settings, demonstrating improved robustness in convection-dominated regimes.

\paragraph{2D Helmholtz.}
For the 2D Helmholtz system, we investigate extrapolation to high-frequency oscillatory regimes parameterized by $(a_1,a_2)$. 
These regimes introduce increasingly complex spatial oscillations that are known to challenge standard PINN optimization. 
Table~\ref{tab:helmholtz_mae_methods_transposed} reports both the mean and variance of the MAE across independent runs to evaluate reconstruction quality and convergence stability. 
The results show that baseline PINN methods experience substantial instability and variance in higher-frequency settings, whereas~\mymethod{} consistently achieves lower error with significantly more stable convergence behavior across both isotropic and anisotropic wave propagation regimes.

\paragraph{2D Navier-Stokes (Forward Problem).}
For the 2D incompressible Navier-Stokes system, we evaluate the forward pressure reconstruction problem across Reynolds numbers $\mathrm{Re}\in\{600,800,1000\}$. 
In this setting, the pressure field is recovered without direct supervision, making optimization particularly challenging in high Reynolds number regimes with complex vortex dynamics. 
Table~\ref{tab:ns_pressure_mae} reports both the mean and variance of the pressure MAE across independent runs to analyze convergence stability in this multi-physics setting. 
The results show that~\mymethod{} achieves substantially lower prediction error and reduced variability compared to existing baselines, indicating more stable and reliable convergence behavior in complex fluid dynamics regimes.
\input{sections/C9.Appendix/tables}

\subsection{Ablation Analysis on 1D Convection and 2D Helmholtz}
\label{appendix:ablation_additional_pdes}
\vspace{-0.5ex}
We extend the ablation analysis of \mymethod{} to additional PDE domains, namely \textbf{1D Convection} and \textbf{2D Helmholtz}. As in the main text, we evaluate the effect of \emph{Invariance Encoding} (IE) and \emph{Gated Layer-wise Optimization} (GLO) using the ablation variants: 
`\mymethod{} w/o (IE,GLO)' (random initialization) and `\mymethod{} w/o GLO' (MAML-style initialization). 
\begin{figure}[ht]
  \centering
  \begin{subfigure}[t]{0.45\textwidth}
    \centering
    \includegraphics[width=0.7\textwidth]{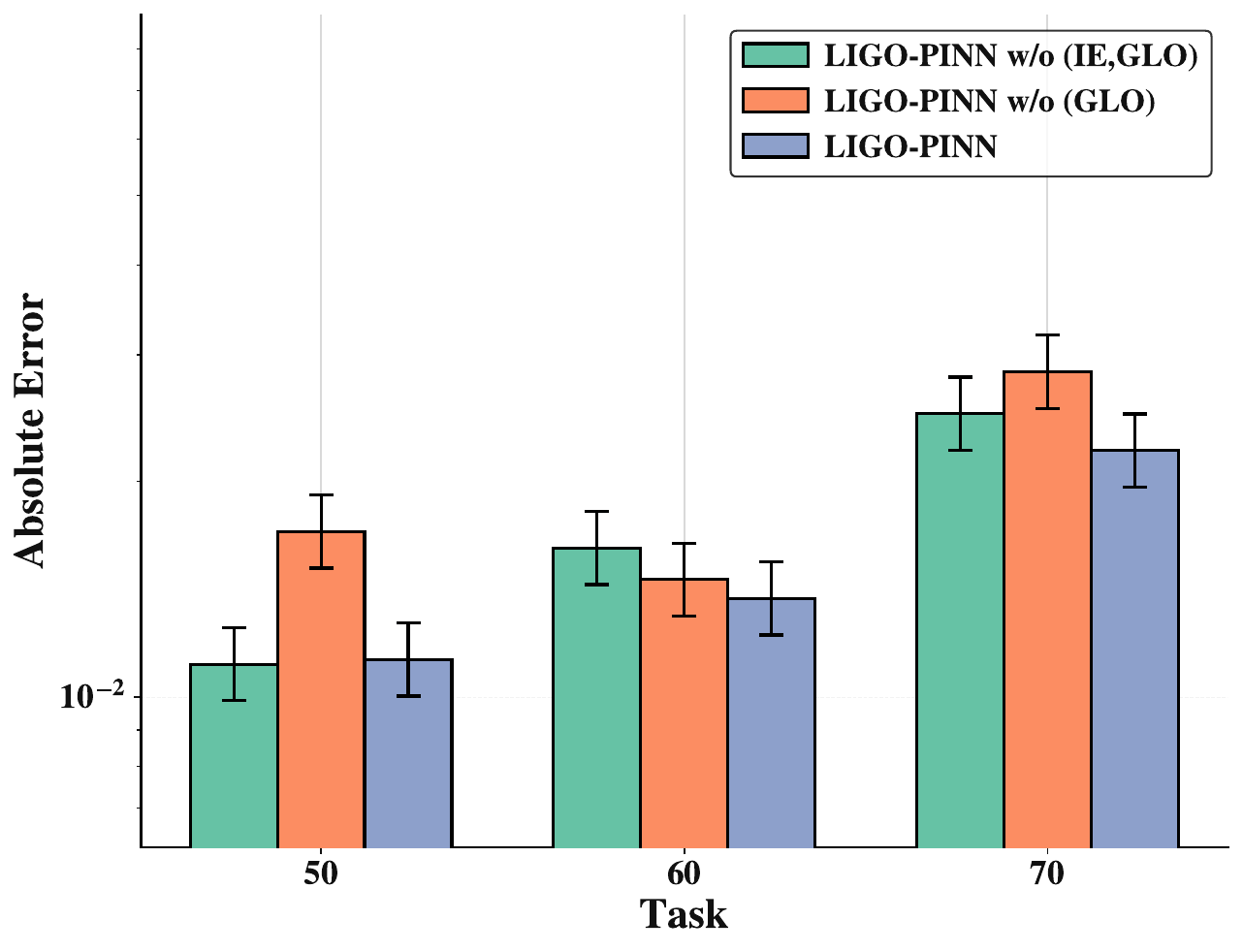}
    \caption{1D Convection (MAE).}
    \label{fig:appendix_convection_empirical}
  \end{subfigure}
  \hfill
  \begin{subfigure}[t]{0.45\textwidth}
    \centering
    \includegraphics[width=0.7\textwidth]{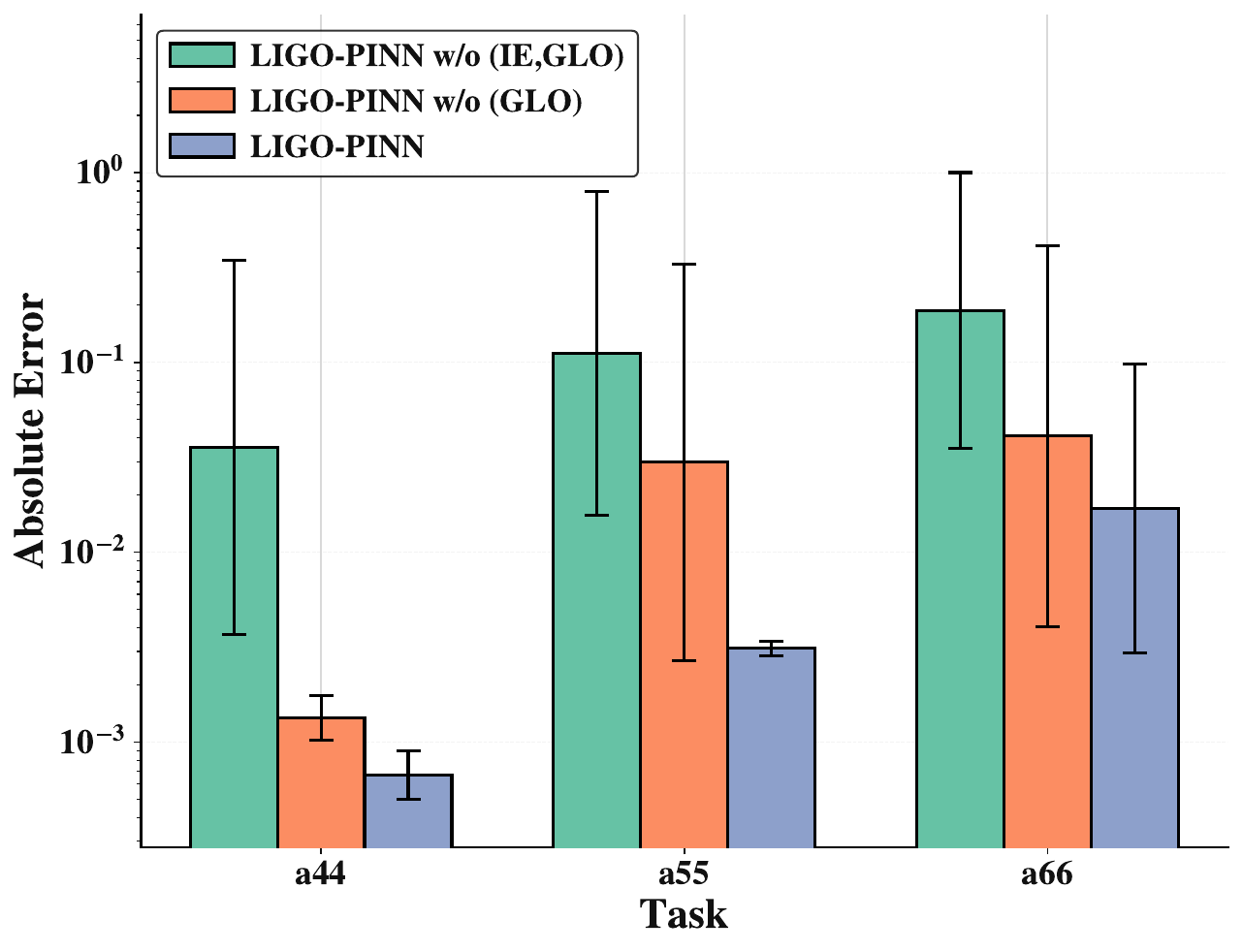}
    \caption{2D Helmholtz (MAE).}
    \label{fig:appendix_helmholtz_empirical}
  \end{subfigure}
  \caption{
  Empirical (MAE) comparison across ablation variants on 1D Convection and 2D Helmholtz. \mymethod{} achieves lower error and reduced performance variability across both domains.
  }
  \label{fig:appendix_empirical}
  \vspace{-2ex}
\end{figure}

\textbf{(i) Performance Trends.}  
Across both PDE domains, random initialization (`w/o (IE,GLO)') results in higher error and larger variance, indicating unstable convergence behavior. Incorporating IE (`w/o GLO') improves performance and reduces variance relative to random initialization, demonstrating the benefit of task-aware initialization. However, the absence of GLO still leads to suboptimal convergence, particularly as task complexity increases. Full \mymethod{}, combining IE and GLO, consistently achieves lower error and tighter error distributions, indicating improved training stability and robustness.

\begin{figure}[!h]
  \centering
  \begin{subfigure}[t]{0.45\textwidth}
    \centering
    \includegraphics[width=0.7\textwidth]{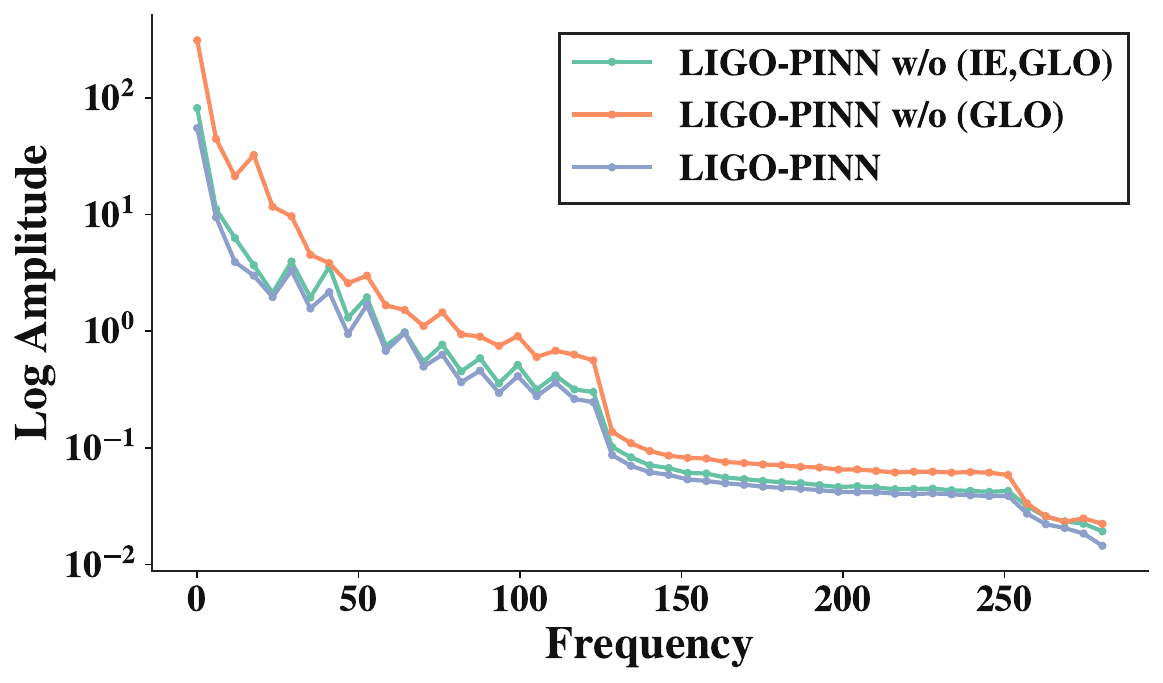}
    \caption{1D Convection ($\beta=70$).}
    \label{fig:appendix_convection_spectral}
  \end{subfigure}
  \hfill
  \begin{subfigure}[t]{0.45\textwidth}
    \centering
    \includegraphics[width=0.7\textwidth]{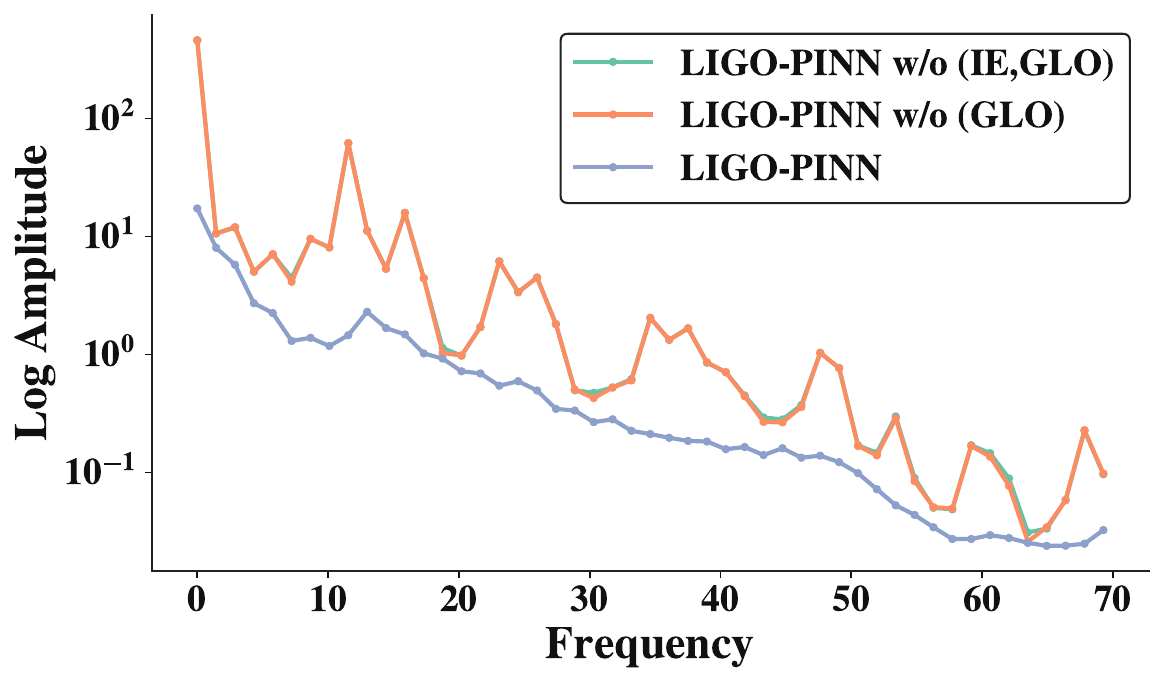}
    \caption{2D Helmholtz ($a_1=6, a_2=6$).}
    \label{fig:appendix_helmholtz_spectral}
  \end{subfigure}
  \caption{
  Spectral distribution of absolute error for \mymethod{} and ablation variants. Lower spectral density indicates better reconstruction of both low- and high-frequency components.
  }
  \label{fig:appendix_spectral}
  \vspace{-2ex}
\end{figure}

\textbf{(ii) Spectral Bias Analysis.}  
We analyze spectral bias by computing the Fourier transform of the absolute error fields. Across both 1D Convection and 2D Helmholtz, ablation variants exhibit higher spectral density, particularly in the high-frequency regime, indicating poor capture of fine-scale features. In contrast, \mymethod{} consistently achieves lower spectral density across both low and high frequencies, demonstrating improved representation of complex solution structures. This confirms that the combination of invariance encoding and gated layer-wise optimization effectively mitigates spectral bias beyond what is achieved by either component alone.

\subsection{Sensitivity Analysis of LIGO-PINN to Number and Range of IE Tasks}
\label{appen:sensitivity-analysis}

We further analyze the sensitivity of LIGO-PINN to two factors during the Invariance Encoding (IE) phase:  
(i) the number of sampled tasks $k$, and (ii) the range of tasks used.  
For fairness, the baseline task ranges in the main paper were selected to align with prior work~\cite{krishnapriyan_characterizing_2021}, which we also compared against through the \methodcurriculumn{} baseline.

\vspace{-1.5ex}
\paragraph{Sensitivity to Number of Tasks $k$.} 
We varied $k \in \{2,3,4\}$ during meta-training, with $k=3$ used in the main paper.  
Evaluation focused on the challenging extrapolation regimes of the 1D convection PDE, $\beta \in \{50,60,70\}$.  
Table~\ref{tab:sensitivity-k} reports the median MAE across random seeds.  
The results show that performance remains relatively stable across different values of $k$, with $k=3$ achieving the best median MAE on $\beta=50$ and $\beta=70$, while all settings remain in a similar range.

\begin{table}[!ht]
\small
\centering
\caption{Sensitivity of LIGO-PINN to the number of sampled tasks $k$ during the IE phase. We report the median mean absolute error (MAE) across random seeds for 1D Convection extrapolation tasks.}
\label{tab:sensitivity-k}
\begin{tabular}{lccc}
\toprule
\textbf{1D Convection ($\beta$)}
& \textbf{$k=2$}
& \textbf{$k=3$}
& \textbf{$k=4$} \\
\midrule
50 & 0.01535 & 0.01285 & 0.01525 \\
60 & 0.01490 & 0.01645 & 0.01600 \\
70 & 0.02725 & 0.01690 & 0.01730 \\
\bottomrule
\end{tabular}
\end{table}

\vspace{-1.5ex}
\paragraph{Sensitivity to Task Range.} 
We further test the robustness of LIGO-PINN by varying the range of easy tasks selected for IE.  
In addition to the main setup LIGO-PINN(5-25), we evaluate LIGO-PINN(10-30) and LIGO-PINN(15-35).  
The evaluation tasks remain the challenging extrapolation regimes $\beta \in \{50,60,70\}$.  
Table~\ref{tab:sensitivity-range} reports the median MAE across random seeds.  
The results show that all LIGO-PINN variants achieve comparable performance across task ranges, suggesting that the method is not overly sensitive to the exact IE task range.

\begin{table}[!ht]
\small
\centering
\caption{Sensitivity of LIGO-PINN to the task range used during the IE phase. We report the median mean absolute error (MAE) across random seeds for 1D Convection extrapolation tasks.}
\label{tab:sensitivity-range}
\begin{tabular}{lccc}
\toprule
\textbf{1D Convection ($\beta$)}
& \textbf{LIGO (5-25)}
& \textbf{LIGO (10-30)}
& \textbf{LIGO (15-35)} \\
\midrule
50 & 0.00535 & 0.01465 & 0.01520 \\
60 & 0.01710 & 0.01505 & 0.02055 \\
70 & 0.01580 & 0.02500 & 0.01350  \\
\bottomrule
\end{tabular}
\end{table}
\vspace{-1.5ex}
\paragraph{Summary.} 
These results suggest that LIGO-PINN is robust to both (i) the number of sampled tasks $k$ and (ii) the range of tasks selected during the IE phase.  
Across the tested settings, median MAE remains consistently low on challenging extrapolation tasks, indicating that the IE phase does not require highly precise tuning of either task count or task range.  
This supports the practical use of domain knowledge to select reasonable IE task ranges without sacrificing downstream extrapolation performance.

%% file: sections/c6.ResultsAnalysis/r4.tex
\subsection{Do \mymethod{} generalize effectively to PDEs defined on complex geometries?}\label{appen:3d_domain_results}
\begin{figure}[!h]
  \centering
   \includegraphics[width=0.9\textwidth]{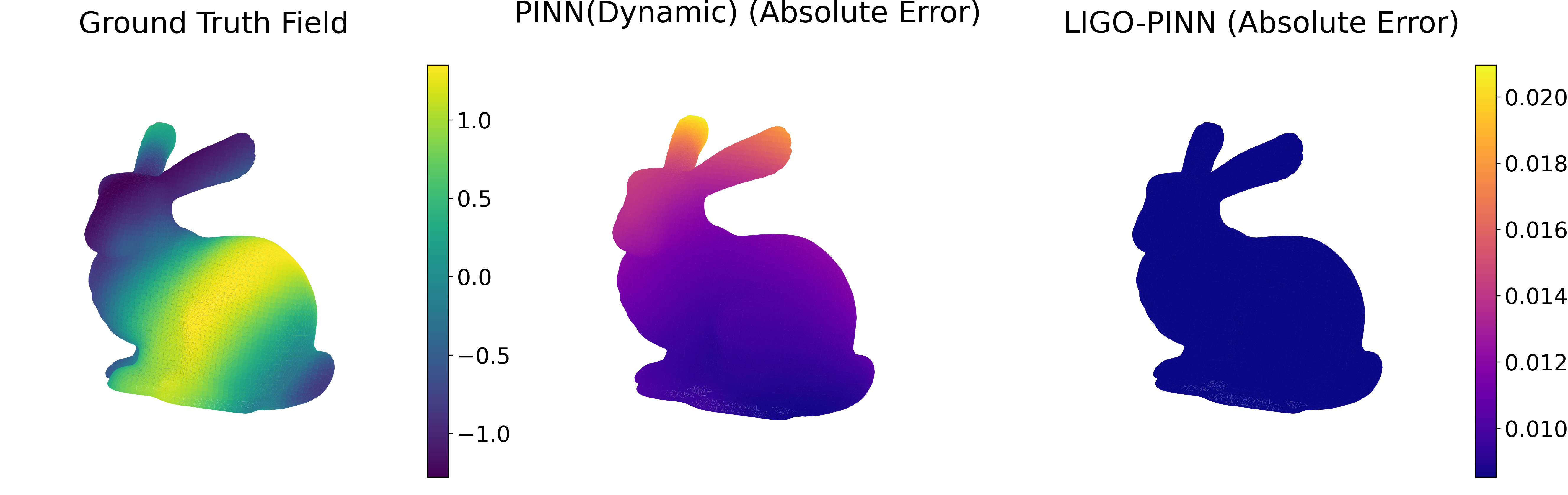}
  \caption{Poisson equation on a bunny surface. Overall, we find that the ~\mymethod{} achieves a ~15\% improvement in MAE over \methodpinnd{} even on complex domains. (a) Ground truth field. (b) LIGO-PINN absolute error. (c) \methodpinnd{} absolute error.}
  \label{fig:poisson_bunny}
\end{figure}
To evaluate whether our method generalizes to PDEs on complex geometries, we compare \mymethod{} with \methodpinnd{} for the screened Poisson equation defined on the Stanford Bunny surface \cite{Turk1994}. We further evaluate both \methodpinnd{} and \mymethod{} under multiple random seed initializations to ensure robustness. Both models are trained for 50k epochs under identical training settings. \mymethod{} achieves a 15\% improvement over \methodpinnd{}. 

%% file: sections/C9.Appendix/AE5.effect_of_initialization.tex
\subsection{Exploring Effect of different initialization}
\label{appen:additional-results-different-init}
Table~\ref{tab:app_heat_extrapolation} compares the extrapolation MAE for PINN-Dynamic under Xavier initialization, PINN with Kaiming initialization, and PINN-Dynamic with LIGO (i.e., nothing but our proposed LIGO-PINN model). 
The learned initialization (i.e., PINN-Dynamic LIGO) achieves consistently lower error than both Xavier and Kaiming baselines as task get progressively harder, indicating that the learned invariance weights provide a stronger starting point for optimization than standard random initializations.

\begin{table}[ht]
\centering
\small
\setlength{\tabcolsep}{4pt}
\caption{Error on 1D convection extrapolation tasks measured by MAE ($\beta \in \{40,50,60,70\}$). Values show mean with standard deviation in parentheses.}
\label{tab:app_heat_extrapolation}
\begin{tabular}{lccc}
\toprule
\textbf{Task}
& \makecell{\textbf{LIGO-PINN}\\ \textit{Xavier}}
& \makecell{\textbf{LIGO-PINN}\\ \textit{Kaiming}}
& \makecell{\textbf{LIGO-PINN}\\ \textit{LIGO}} \\
\midrule

40
& 0.0130 (0.0065)
& 0.0158 (0.0029)
& \textbf{0.0072 (0.0012)} \\

50
& 0.0120 (0.0044)
& 0.0176 (0.0023)
& \textbf{0.0119 (0.0032)} \\

60
& 0.0162 (0.0019)
& 0.0210 (0.0057)
& \textbf{0.0137 (0.0024)} \\

70
& 0.0256 (0.0070)
& 0.0708 (0.0644)
& \textbf{0.0179 (0.0050)} \\

\midrule
Avg
& 0.0167
& 0.0313
& \textbf{0.0127} \\
\bottomrule
\end{tabular}
\end{table}




\par\noindent
Layer-wise Parameter Distribution Analysis.
To characterize how the learned initialization modifies the PINN parameter landscape, we compare the layer-wise weight distributions of LIGO-PINN against layer-wise weight distributions of a randomly initialized PINN. We report (Table.~\ref{tab:stats_kde}) two statistics for the first, middle, and final layers across all PDEs namely: (i) the absolute difference between the variance of weight distributions per layer $\Delta = |\mathrm{Var}(\text{LIGO-PINN}) $-$ \mathrm{Var}(\text{Random PINN})|$, and (ii) the 1-Wasserstein distance, which measures distributional similarity.

\begin{table}[ht]
\small
\centering
\caption{Layer-wise statistics across PDEs. Variance $\Delta$ captures the absolute scale difference and Wasserstein distance quantifies distributional similarity.}
\label{tab:stats_kde}
\begin{tabular}{
    l
    |r|r   
    |r|r   
    |r|r   
}
\toprule
\multirow{2}{*}{\textbf{Layer}}
& \multicolumn{2}{c|}{\textbf{1D Convection}}
& \multicolumn{2}{c|}{\textbf{2D Helmholtz}}
& \multicolumn{2}{c}{\textbf{2D Navier--Stokes}}
\\
& \textbf{Variance $\Delta$} & \textbf{Wasserstein}
& \textbf{Variance $\Delta$} & \textbf{Wasserstein}
& \textbf{Variance $\Delta$} & \textbf{Wasserstein}
\\
\midrule
First Layer   & 0.058103 & 0.108291 & 0.013771 & 0.038730 & 0.004854 & 0.039123 \\
Middle Layer  & 0.009142 & 0.024560 & 0.002109 & 0.006187 & 0.001054 & 0.004970 \\
Last Layer    & 0.000264 & 0.002663 & 0.000123 & 0.002950 & 0.000459 & 0.004678 \\
\bottomrule
\end{tabular}
\end{table}
In Table~\ref{tab:stats_kde}, we find an interesting behavior where the results show a consistent pattern across systems. Specifically, we found a stark distributional dissimilarity (between randomly initialized PINNs and PINNs initialized with learned weights) in shallower layers with increased distributional similarity in deeper layers. This difference in distributional similarity can be attributed to a higher variance of weights in shallower layers of \mymethod{}, compared to variance of weights in the corresponding layers of randomly initialized PINNs. Previous work has shown that neural networks with higher variance of weights in shallower layers have increased expressivity.\cite{yang2017mean}, thereby allowing us to infer that \mymethod{} with learned weights offer greater expressive power (and thereby larger learning capacity) compared to randomly initialized PINNs, also contributing to their superior performance in challenging PDE domains.
\hide{The first layer exhibits the largest discrepancy, indicating that the learned initialization primarily reshapes how inputs are encoded. The mismatch decreases in deeper layers, and the final-layer distributions are nearly identical. This suggests that the meta-learned initialization focuses on early feature processing while preserving the output-layer structure.}

%% file: sections/C9.Appendix/tables.tex
\begin{table}[h]
\centering
\scriptsize
\setlength{\tabcolsep}{3pt}
\caption{Mean Absolute Error (MAE) on 1D convection for extrapolation tasks 
($\beta \in \{40, 50, 60, 70, 80\}$). 
Values report the mean MAE across 5 random seeds, with corresponding standard deviation in parentheses.}
\label{tab:1dconv_mean_stdev_extrapolation}
\begin{tabular}{lccccc}
\toprule
\textbf{Method} & \textbf{40} & \textbf{50} & \textbf{60} & \textbf{70} & \textbf{80} \\
\midrule

\methodpinnf
& 0.7433 (0.1522)
& 0.7337 (0.2567)
& 0.8720 (0.1020)
& 0.7671 (0.1478)
& 0.8253 (0.1761) \\

\methodpinnd
& 0.0095(0.0050)
& 0.0148(0.0050)
& 0.0881(0.1534)
& 0.1597(0.1988)
& 0.1699(0.2072) \\

\methodcurriculumn
& 0.0801 (0.0074)
& 0.2593 (0.2183)
& 0.3215 (0.0845)
& 0.3726 (0.0272)
& 0.4072 (0.0248) \\

\methodevo
& 0.0162 (0.0125)
& 0.1477 (0.1823)
& 0.3557 (0.0138)
& 0.4327 (0.0334)
& 0.4151 (0.0166) \\

\methodmetapde
& 0.1588 (0.1405)
& 0.2112 (0.1270)
& 0.3757 (0.1548)
& 0.3570 (0.0214)
& 0.5167 (0.1058) \\

\methodhyperlr
& 0.4008 (0.0250)
& 0.4584 (0.0309)
& 0.4912 (0.0261)
& 0.5359 (0.0248)
& 0.5574 (0.0253) \\

\methodmaml
& 0.0126 (0.0030)
& 0.0181 (0.0069)
& 0.0149 (0.0035)
& 0.0377 (0.0365)
& 0.0524 (0.0330) \\

\textbf{\mymethod{} (Ours)}
& \textbf{0.0072(0.0012)}
& \textbf{0.0119(0.0032)}
& \textbf{0.0137(0.0024)}
& \textbf{0.0179(0.0050)}
& \textbf{0.0871(0.1466)} \\

\bottomrule
\end{tabular}
\end{table}

\begin{table}[h]
\centering
\scriptsize
\setlength{\tabcolsep}{3pt}
\caption{Mean Absolute Error (MAE) on the 2D Helmholtz PDE across varying $a_{xy}$ in the extrapolation setting. 
Values report the mean MAE across random seeds, with corresponding standard deviation reported in parentheses.}
\label{tab:helmholtz_mae_methods_transposed}
\begin{tabular}{lcccccc}
\toprule
\textbf{Method} & $a_{44}$ & $a_{45}$ & $a_{55}$ & $a_{46}$ & $a_{56}$ & $a_{66}$ \\
\midrule

\methodpinnf
& 0.1135 (0.0757)
& 0.2276 (0.0993)
& 0.5931 (0.3411)
& 0.3447 (0.2861)
& 0.7396 (0.3906)
& 1.2283 (0.5588) \\

\methodpinnd
& 0.1615(0.2152)
& 0.3956(0.4806)
& 0.3186(0.1753)
& 0.3185(0.1759)
& 0.8444(1.0390)
& 0.4784(0.2969) \\

\methodcurriculumn
& 0.3080 (0.4643)
& 0.1630 (0.1448)
& 0.8223 (1.3994)
& 0.1331 (0.1515)
& 0.4177 (0.1481)
& 0.6520 (0.5460) \\

\methodevo
& 0.4910 (0.1544)
& 0.4192 (0.0491)
& 0.5143 (0.1766)
& 0.4786 (0.1512)
& 0.4550 (0.0669)
& 0.3968 (0.0003) \\

\methodmetapde
& 0.2473 (0.1541)
& 0.4736 (0.3218)
& 0.5982 (0.3420)
& 0.4954 (0.3495)
& 0.4077 (0.0092)
& 0.4329 (0.0578) \\

\methodhyperlr
& 0.0613 (0.0317)
& 0.0435 (0.0140)
& 0.1599 (0.1437)
& 0.1570 (0.1007)
& 0.7585 (0.8403)
& 2.9437 (0.8651) \\

\methodmaml
& 0.0014 (0.0004)
& 0.1604 (0.2161)
& 0.1657 (0.2195)
& 0.1617 (0.2148)
& 0.1625 (0.2141)
& 0.2134 (0.2951) \\

\textbf{\mymethod{} (Ours)}
& \textbf{0.0007 (0.0002)}
& \textbf{0.0017 (0.0004)}
& \textbf{0.0031 (0.0003)}
& \textbf{0.0032 (0.0007)}
& \textbf{0.0048 (0.0010)}
& \textbf{0.0800 (0.1600)} \\

\bottomrule
\end{tabular}
\end{table}

\begin{table}[h]
\small
\centering
\caption{Pressure error (MAE) for 2D Navier--Stokes across Reynolds numbers. 
Values are mean with standard deviation in parentheses.}
\label{tab:ns_pressure_mae}
\begin{tabular}{lcccc}
\toprule
\textbf{Method} & \textbf{600} & \textbf{800} & \textbf{1000} & \textbf{Avg} \\
\midrule

\methodpinnf
& 0.0183 (0.0135)
& 0.1934 (0.3643)
& 0.0257 (0.0188)
& 0.0791 \\

\methodpinnd
& 0.0239 (0.0314)
& 0.0958 (0.2027)
& 0.0353 (0.0663)
& 0.0517 \\

\methodcurriculumn
& 0.3011 (0.3055)
& 0.2597 (0.2683)
& 0.2309 (0.2555)
& 0.2639 \\

\methodevo
& 0.3519 (0.2572)
& 0.4966 (0.3249)
& 0.2620 (0.2021)
& 0.3702 \\

\methodmetapde
& 0.3172 (0.0400)
& 0.2997 (0.0262)
& 0.2812 (0.0342)
& 0.2994 \\

\methodhyperlr
& 0.2155 (0.0594)
& 0.1253 (0.0022)
& 0.1910 (0.0273)
& 0.1773 \\

\methodmaml
& 0.1510 (0.1859)
& 0.1941 (0.1773)
& 0.0288 (0.0302)
& 0.1246 \\

\textbf{\mymethod{} (Ours)}
& \textbf{0.0164 (0.0187)}
& \textbf{0.0065 (0.0011)}
& \textbf{0.0058 (0.0001)}
& \textbf{0.0096} \\

\bottomrule
\end{tabular}
\end{table}